%% file: main.tex
\documentclass[10pt,twocolumn,letterpaper]{article}

\usepackage[pagenumbers]{iccv} 

\input{preamble}

\definecolor{iccvblue}{rgb}{0.21,0.49,0.74}
\usepackage[pagebackref,breaklinks,colorlinks,allcolors=iccvblue]{hyperref}

\title{Is there anything left? Measuring semantic residuals of objects removed from 3D Gaussian Splatting}

\author{
Simona Kocour$^{1,2}$\\
\and
Assia Benbihi$^2$\\
\and
Aikaterini Adam$^3$\\
\and
Torsten Sattler$^2$\\
\and
$^1$Faculty of Electrical Engineering, Czech Technical University in Prague \\
$^2$Czech Institute of Informatics, Robotics and Cybernetics, Czech Technical University in Prague\\
$^3$Archimedes/Athena RC\\
{\tt\small simona.kocour@cvut.cz}
}

\begin{document}
\maketitle

\begin{abstract}
Searching in and editing 3D scenes has become extremely intuitive with trainable scene representations that allow linking human concepts to elements in the scene.
These operations are often evaluated on the basis of how accurately the searched element is segmented or extracted from the scene.
In this paper, we address the inverse problem, that is, how much of the searched element remains in the scene after it is removed.
This question is particularly important in the context of privacy-preserving mapping when a user reconstructs a 3D scene and wants to remove private elements before sharing the map.
To the best of our knowledge, this is the first work to address this question.
To answer this, we propose a quantitative evaluation that measures whether 
a removal operation 
leaves object residuals that can be reasoned over.
The scene is not private when such residuals are present.
Experiments on state-of-the-art scene representations show that the proposed metrics are meaningful
and consistent with the user study that we also present.
We also propose a method to refine the removal based on spatial and semantic consistency.
The evaluation framework and data are available at \url{https://github.com/simonakocour/anything_left}
\end{abstract}

\section{Introduction}

Trainable scene representations, such as neural radiance fields (NeRFs)~\cite{mildenhall2020nerf,barron2022mip,reiser2021kilonerf,muller2022instant,chen2024lara,kulhanek2023tetra,martin2021nerf} or 3D Gaussians~\cite{kerbl3Dgaussians,lin2024vastgaussian,yu2024gaussian,zhang2024gaussian,kulhanek2024wildgaussians,chen2024mvsplat,wang2024freesplat}, enable photorealistic 3D reconstructions from images.
They are gaining tremendous popularity thanks to the high quality of the reconstructed geometry and the realism of the renderings.

These learnable scene representations can be easily extended to include features that carry semantic meaning~\cite{kerr2023lerf,shi2024language,gaussian_grouping,wuopengaussian,zhou2024feature,hu2024semantic,jain2024gaussiancut}. 
This, in turn, creates a link between parts of the scene and human understandable concepts in general and human language in particular. 
This link can then be used for intuitive search in 3D scene representation by prompting~\cite{peng2023openscene,huang2024segment3d,takmaz2025search3d,liang2024supergseg,koch2024relationfield}, \eg, asking 'find the remote control'.
Similarly, it enables intuitive editing operations~\cite{gaussian_grouping,zhou2024feature,chen2024dge,gu2025egolifter,choi2024click}, \eg, by asking to 'remove the red armchair in the living room'. 

The growing availability of learning-based 3D reconstruction software accessible to nonexpert users~\cite{scaniverse,nerfstudio,Yu2022SDFStudio,capturereality,polycam,ye2024gsplatopensourcelibrarygaussian,postshot,lumaAi}, coupled with intuitive edit operations based on natural language~\cite{radford2021learning,achiam2023gpt,schuhmann2022laion}, opens up exciting possibilities:
using data casually captured by a smart phone~\cite{scaniverse,polycam,realityscan,lumaAi}, a user can easily create and edit photorealistic 3D models.  
In particular, they can remove parts of the scene that they consider private, \eg, photos, documents, types of decoration, \etc, before sharing these maps, \eg, before uploading them to online services for refurbishing\footnote{As an example, IKEA provides an app that allows users to scan rooms.
The captured data is uploaded to IKEA's servers.
Hence, IKEA recommends to physically remove private parts before scanning. In contrast, the systems envisioned in this paper allow to perform the removal virtually after the scan, which is more practical and less physically demanding.}.
The privacy implications of sharing 3D maps naturally lead to questions on how well existing approaches for intuitive edit operations perform. 

\begin{figure*}[t]
    \centering
    \includegraphics[width=\linewidth]{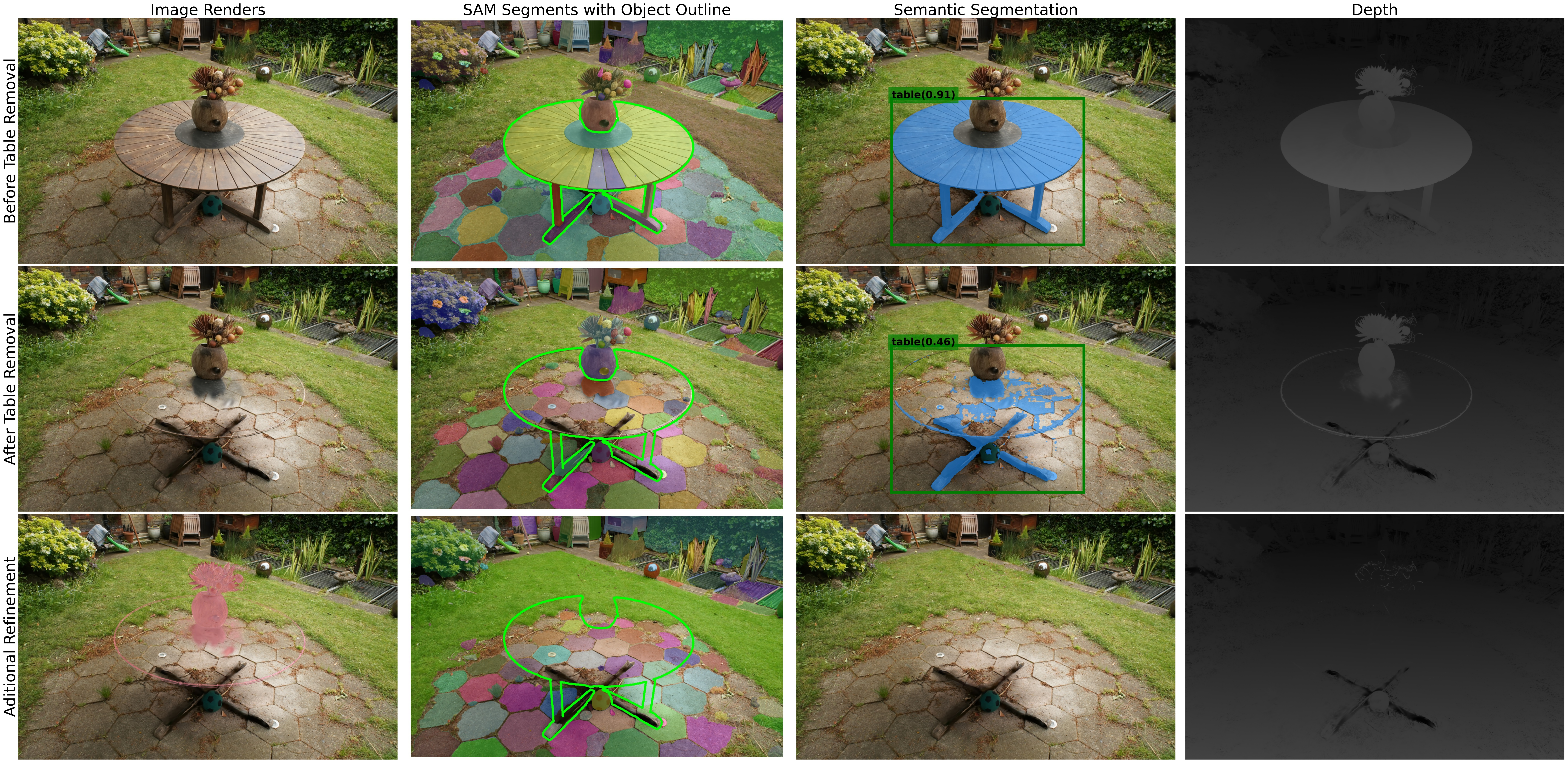}
    \caption{
    \textbf{Is there any table left after it is removed from the scene?}
    When there remain residuals of the object, and they can be reasoned over (left column, mid-row), the object removal is imperfect.
    This is what the proposed evaluation measures based on whether off-the-shelf semantic models can still segment the object after the removal.
    We also measure the presence of residuals in 3D with depth (right column, mid-row).
    We derive an optimization to refine the removal based on spatial and semantic consistency (highlighted in pink the bottom-left image).
    Top-Bottom: Rendering from the original 3DGS~\cite{kerbl3Dgaussians} scene, rendering after removing the table from the scene, rendering after removal and refinement.
    Left-Right: RGB renderings, SAM~\cite{kirillov2023segany} masks overlay with pseudo-ground-truth object outline, GroundedSAM~\cite{kirillov2023segany,liu2023grounding,ren2024grounded} overlay, depth renderings.
    }
    \label{fig:teaser}
\end{figure*}

This paper is dedicated to this privacy aspect of (language-based) editing of trainable scene representations.
We focus on removing objects from scenes and investigating whether state-of-the-art removal methods leave residuals that enable us to reason about what content was removed.
Contrary to previous works~\cite{cen2023saga,choi2025click,mirzaei2023spin} that focus on foreground / background segmentation, we are interested in whether the residuals of the objects remain in the scene after removal and whether the residuals can be reasoned over.
To the best of our knowledge, this is the first work that considers this aspect of trainable scene representations. 
Hence, this paper focuses on introducing an evaluation framework 
to quantitatively and qualitatively measure how well the current state-of-the-art methods remove objects from the scene.
We propose 
metrics that measure how well an object is recognized after removal using off-the-shelf semantic segmentation at various granularity levels (object semantics, instances, and object parts) and 3D perception with depth.
Experiments on indoor and outdoor scenes show that the proposed metrics are consistent in their ranking of the evaluated methods.
Also, their complementarity makes the proposed evaluation robust to errors in the semantic segmentations.
We complete the quantitative results with qualitative insights drawn from a user study with human volunteers that evaluate the removal methods.
The human conclusions are mostly consistent with the quantitative metrics we define.
We also derive a graph-based optimization to refine the removal that leverages the tight coupling between the geometry and the semantics of the scene.
Results show that the refinement either preserves or improves the quality of the object removal.
In summary, our contributions are: 
i) we propose an evaluation that measures how well scene removal operations remove objects in the context of privacy. 
To the best of our knowledge, this is the first work that considers this aspect of trainable scene representations; 
ii) we define 
quantitative metrics to enable this evaluation
and demonstrate their soundness on state-of-the-art trainable scene representations; 
iii) we complete our quantitative evaluation with a user study that draws insights consistent with the metrics; 
iv) we define a graph-based optimization that refines the object removal by enforcing spatial and semantic consistency in the removal.
The code and the data used in the evaluation are available at \url{https://github.com/simonakocour/anything_left}

\section{Related Work}

\PAR{3D reconstruction from images} builds 3D models that capture the scene's geometry and appearance. 
Popular scene representations are point clouds~\cite{schonberger2016structure,yang2023sam3d,huang2024segment3d,peng2023openscene,liu2024uni3d,yin2024sai3d}, meshes~\cite{schoenberger2016mvs,furukawa2009accurate,lazebnik2001computing,kundu2020virtual}, and, more recently, Neural Radiance Fields (NeRFs)~\cite{mildenhall2020nerf,barron2022mip,reiser2021kilonerf,muller2022instant,chen2024lara,kulhanek2023tetra,martin2021nerf} and 3D Gaussian Splatting (3DGS)~\cite{kerbl3Dgaussians,lin2024vastgaussian,yu2024gaussian,zhang2024gaussian,kulhanek2024wildgaussians,chen2024mvsplat,wang2024freesplat}.

NeRFs represent the scene implicitly with a colored volumetric field:  
for each 3D point in the space, an MLP outputs a volumetric density and a view-dependent color value.
NeRFs are optimized for novel-view synthesis, \ie, so that a rendering of the field from a given view is similar to the original image from that view.
In 3DGS~\cite{kerbl3Dgaussians}, the scene is represented explicitly with a set of 3D Gaussians with learnable parameters (positions, orientation, scale, opacity, view-dependent color).
Again, the parameters are optimized so that splatting, \ie rendering, the Gaussians generates a rendering that is similar to the original view.

\PAR{Linking 3D reconstructions and semantics.} 
NeRFs and 3DGS can easily be extended to embed semantic features that can be prompted via natural language~\cite{kerr2023lerf,tschernezki2022neural}, pixel locations~\cite{kerr2023lerf,tschernezki2022neural} or semantic labels~\cite{gaussian_grouping}.
The prompt
allows humans to search for elements in the 3D reconstruction then edit that location. 
The type of prompt and the edition operation depend on the 3D representation and the embedded semantic features.

In NeRFs, the MLP is extended to output semantic features that are rendered in a differential manner (as for the color) and supervised with off-the-shelf 2D semantic features.
Examples are NeRFs extended with text features~\cite{wang2022clip,mirzaei2022laterf} like CLIP~\cite{radford2021learning}, text and semantic features~\cite{kerr2023lerf,kobayashi2022decomposing}, and unsupervised features~\cite{tschernezki2022neural} like DINO~\cite{caron2021emerging,oquab2024dinov2}.
Prompting for an element in the reconstruction amounts to finding the features in the NeRF field that are the most similar to the prompt's features.
The feature field can also be trained to render segmentation masks consistent with 2D  masks~\cite{zhi2021place,nesf,cen2023segment,liu2024sanerf} derived with semantic models~\cite{chen2017deeplab,graham2017submanifold,lilanguage}, panoptic models~\cite{siddiqui2023panoptic,bhalgat2024contrastive}, foundation models~\cite{ravi2024sam2,kirillov2023segany,zou2023segment,li2024segment,liu2022open}, or user annotations~\cite{mirzaei2023spin}.
The prompt search then looks for the features associated with a given semantic label.

Although these methods perform well in locating prompted elements in the reconstruction, the implicit nature of NeRFs makes it complex to associate a location in the reconstruction with a set of abstract MLP parameters to edit.
Hence, editing the reconstruction is not as straightforward as in 3DGS~\cite{kerbl3Dgaussians} representations that are explicit: removing a prompted location amounts to deleting the 3D Gaussians at that location.
3DGS can also be embedded with semantic features as easily as NeRFs with semantic models~\cite{cen2023saga,choi2025click,jain2024gaussiancut,wu2024opengaussian,gu2025egolifter,zhou2024feature}, text features~\cite{shi2024language,liao2024clip,qin2024langsplat}, and unsupervised features~\cite{zuo2024fmgs}.
This motivates this paper to focus on 3D reconstructions represented with 3DGS~\cite{kerbl3Dgaussians}.

\PAR{Evaluating 3D reconstruction approaches.} 
3D reconstructions are evaluated based on the accuracy of the 3D geometry and whether the color renderings and the semantic (features) renderings are similar to those in the training views.

Often, scene operations, such as editing and removal, are reported only as illustrative examples with qualitative results~\cite{gaussian_grouping,zhou2024feature,gu2025egolifter,wu2024opengaussian,jain2024gaussiancut,choi2025click}.
Some works~\cite{cen2023saga,choi2025click,mirzaei2023spin} report a quantitative evaluation of `background' removal by comparing the renderings of the searched element against its appearance in the original view.
This paper addresses a different problem, \ie, are there residuals of the removed object in the scene and if so, can they still be recognized or reasoned over?
We propose an evaluation framework to answer this question in a quantitative and automatic manner.
To the best of our knowledge, there is no previous work that addresses such a question.

\PAR{Privacy challenges.}
The use of extensive public data in the recent scientific breakthroughs~\cite{rombach2022high,saharia2022photorealistic,brooks2024video,achiam2023gpt} has drawn attention to the protection of user data in the research community~\cite{raina2023egoblur,speciale2019privacy,pittaluga2019revealing,chelani2023privacy,moon2024raycloud,nasr2023effectively}, in companies~\cite{rubinstein2013privacy,grynbaum2023times}, and governments~\cite{illman2019california,voigt2017eu}.

This challenge will grow as the deployment in households of new types of sensors, \eg, Augmented Reality / Virtual Reality (AR / VR) glasses~\cite{spectacles,aria,xreal}, and autonomous systems will become the norm.
An efficient way to make systems privacy-preserving is to consume data that has already been made privacy-preserving by the user.
One relevant line of work proposes to anonymize the images~\cite{liu2024infusion,weder2023removing} with inpainting before the scene reconstruction rather than editing the reconstruction later.

However, this method is more computationally complex: it involves editing many images as opposed to a single reconstruction and can introduce artifacts in the image that reduce the quality of the reconstruction.
Hence, operating on the 3D model offers privacy at a reasonable computational cost and better reconstruction quality.

\section{Metrics Definition}

We evaluate whether removal operations in 3DGS~\cite{kerbl3Dgaussians} leave information about the removed content.
The proposed evaluation is conducted in the context of privacy considerations raised when sharing the scene representations.
Within this context, an element is considered private if it can not be identified~\cite{illman2019california,voigt2017eu,raina2023egoblur}.

This section defines the metrics that evaluate whether elements can be identified after removal.

\subsection{Semantic Object Recognition}
\label{sec:method_iou_drop}

Semantic segmentation identifies objects in the scene by classifying each pixel into semantic categories~\cite {chen2017deeplab,graham2017submanifold,lilanguage}.
Here, it is used to assess whether an object can be identified by rendering the scene from multiple views and measuring the performance of the segmentation model on those renderings.
Comparing the segmentation performance before and after removal provides information on the removal quality.
A drop in performance indicates that the object is removed.

We thus define the semantic recognition metric as the segmentation's performance gap on the renderings before and after removal.

The semantic segmentation is evaluated with the standard \gls{iou}~\cite{chen2017deeplab,badrinarayanan2017segnet} that measures how well the estimated semantic mask overlaps with the ground-truth mask. 

We compute IoU$_{\text{pre}}$ and IoU$_{\text{post}}$ on the masks estimated on the renderings before and after removal, respectively.
The \gls{iou} drop IoU$_{\text{drop}}$ is defined as:
\begin{equation}
    \text{IoU}_{\text{drop}}^{\uparrow} = \text{IoU}_{\text{pre}} - \text{IoU}_{\text{post}}
\end{equation}
IoU$_{drop}$ ranges between 1 and -1 and the higher, the better the object is removed, as indicated by the $\uparrow$.

Low absolute value of the $\text{IoU}_{\text{drop}}$ implies that $\text{IoU}_{\text{post}} = \text{IoU}_{\text{pre}}$, which can be interpreted in two ways.
(1) Both $\text{IoU}_{\text{post,pre}}$ are high, \ie, the segmentation model identifies the object both before and after removal, so the removal failed.
(2) When both IoUs are low, no conclusion can be drawn since the segmentation model could not segment the object even in the original image.
Thus, there is no performance drop to measure.

To address the ambiguity between these two interpretations, we complete this metric with another semantic metric defined in the next section.

Still, $\text{IoU}_{\text{drop}}$ is relevant on its own to signal an issue with the removal or the evaluation, which can be useful in systems with a man in the loop, \eg, active labeling.

In the experiments, we also report a more intuitive metric, the performance of the segmentation after removal, and analyse its correlation $\text{IoU}_{\text{drop}}$.
We define the accuracy $\text{acc}_{\text{seg}}$ as the ratio of test images in which the semantic element is not recognized anymore.

The element is not recognized if IoU$_{\text{post}}$ is smaller than a given threshold $\xi_{\text{IoU}}$, and we report this metric over multiple thresholds.
$\text{acc}_{\text{seg}}^{\uparrow}$ ranges between $0$ and $1$ and the higher $\text{acc}_{\text{seg}}$, the better the object removal, as indicated by the $\uparrow$.
\begin{equation}
    \text{acc}_{\text{seg},\xi_{\text{IoU}}}^{\uparrow} = \frac{\| \text{\# images with } \text{IoU}_{\text{post}} < \xi_{\text{IoU}} \|}{\| \text{\# images } \|}
\end{equation}

\begin{figure}
    \centering
    \includegraphics[width=\linewidth]{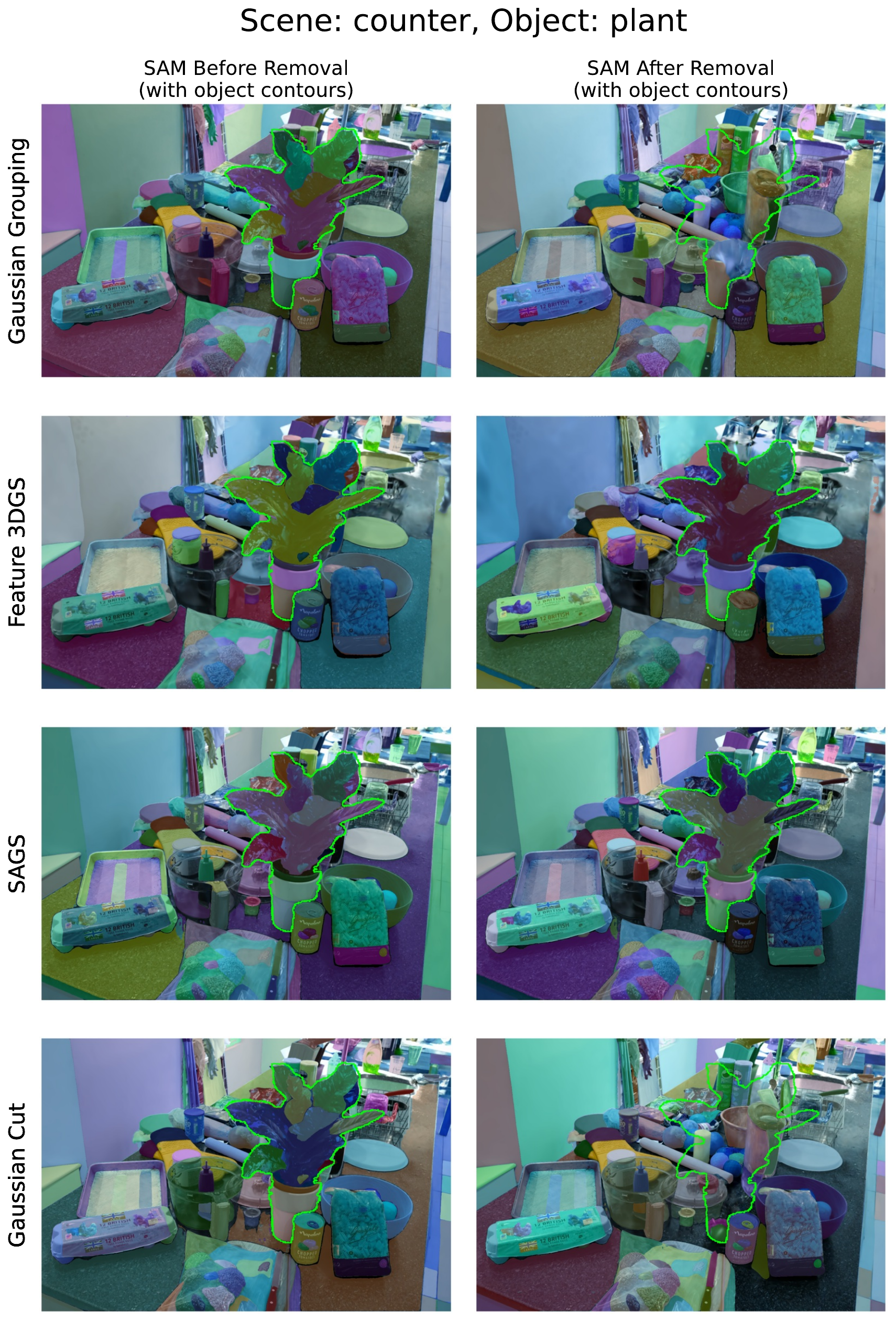}
    \caption{\textbf{SAM~\cite{kirillov2023segany,ravi2024sam2} masks before and after object removal.}
    Left-right: before and after removal. Rows: The evaluated methods.
    When the object (green outline) is removed (row 1 and 4), the SAM masks change, which is the signal we exploit to evaluate the removal: larger changes indicate better removal.
    }
    \label{fig:sam}
\end{figure}

\subsection{Anything Recognition}

\begin{figure*}
    \centering
    \includegraphics[width=\linewidth]{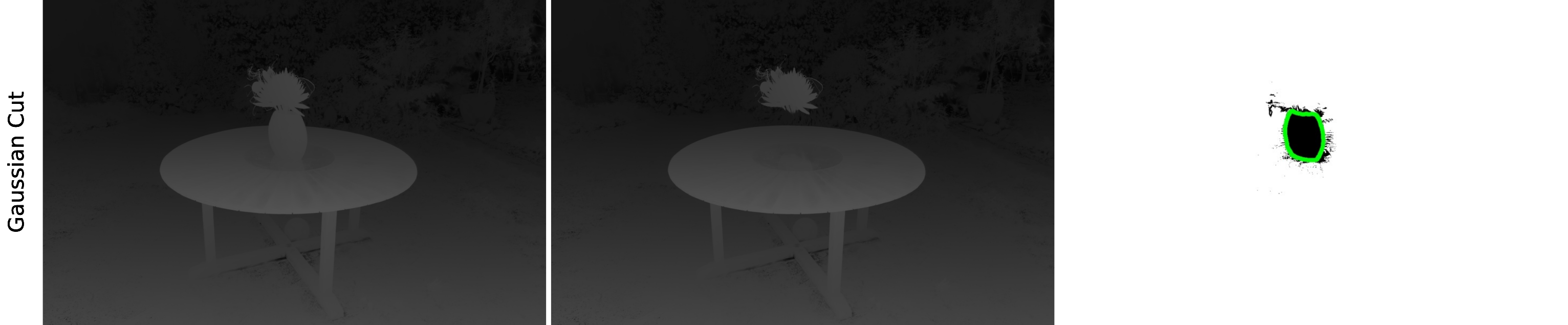}
    \includegraphics[width=\linewidth]{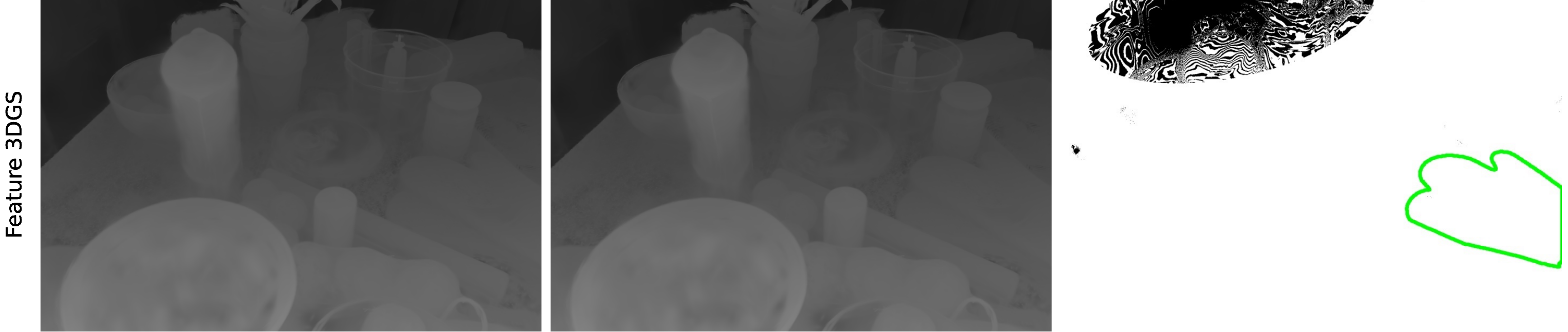}
    \caption{\textbf{Depth Changes before and after removal.}
    Left to right: original rendered depth, rendered depth after removal, thresholded depth difference and pseudo-ground-truth outline of the object to be removed in green.
    Top: GaussianCut~\cite{jain2024gaussiancut} shows precise and localized changes in the depth maps, indicating accurate object removal.
    Down: Feature3DGS~\cite{zhou2024feature} fails to remove the object and the difference at the object location remains unchanged. 
    }
    \label{fig:depth}
\end{figure*}

The previous sections defined the $\text{IoU}_{\text{drop}}$ that can be interpreted in two ways when it is low.
To address such ambiguity, we complement the IoU$_{\text{drop}}$ with a second semantic metric based on finer segmentations, \ie, object parts or instances instead of semantic categories.
These segmentations are derived with the foundational \gls{sam}~\cite{kirillov2023segany,ravi2024sam2},

a prompt-based segmentation model that can be prompted with image locations, bounding boxes, or texts.

The model can also operate without prompts, which results in a set of masks that cover all the semantic elements in the image (See~\cref{fig:sam}).

In this evaluation, we assess whether an element has been removed by comparing the \gls{sam}~\cite{kirillov2023segany,ravi2024sam2} masks of the scene renderings before and after removal.
When the object is removed or partially removed, SAM segments what is behind the object so the masks should change (See rows 1 and 4 in~\cref{fig:sam}).

We next define $\text{sim}_{\text{SAM}}$ that measures the similarity between two sets of SAM~\cite{kirillov2023segany,ravi2024sam2} masks based on how well they spatially overlap.

We first match the masks that overlap the most between the two sets.
Then $\text{sim}_{\text{SAM}}$ is the average overlap between the mask matches.
We enforce a 1-to-1 matching, \ie, a mask in one set is matched to at most one mask in the other.
We do so by defining that two masks match if they overlay, and if one mask gets matched to more than one, we keep the match that leads to the maximum overlay over all matches.
This is derived by solving an assignment problem that maximizes the overlay over all matches with the Hungarian algorithm~\cite{munkres1957algorithms}.

More formally, let $A=(a_i)_{i\in[1,N]}$ and $B=(b_j)_{j\in[1,M]}$ be the sets of \sam masks before and after object removal,

and $(a_k, b_k)_{k=1,K}$ be the $K$ matching masks.
The similarity between the two sets of SAM masks $A$ and $B$ is:
\begin{equation}
    \text{sim}_{\text{SAM}}^{\downarrow} = \frac{\sum_{k=1}^{K} \text{IoU}(a_k, b_k)}{\text{max}(N,M)}
\end{equation}

$\text{sim}_{\text{SAM}}^{\downarrow}$ lies in $[0,1]$ and the lower the score, the less similar the masks, hence the better the removal, as indicated by the $\downarrow$.

Note that we normalize the score with the highest number of masks $\text{max}(N, M)$ instead of the number of mask matches $K$.
We do so to account not only for the difference in overlay (in the numerator) but also for the difference in the number of masks.
To reflect the changes related to the removed object, the metric is derived only over the masks that overlay with the object with an IoU of at least $0.1$.

\subsection{Spatial Recognition}

We complete the previous metrics with one that depends only on the 3D scene before and after removal, hence increasing the robustness of the evaluation against possible errors in the segmentations.

Inspired by recent works in scene change detection~\cite{adam2022objects}, we measure how well an object is removed based on the changes in the rendered depths before and after removal: a strong change in the depth maps indicates a change in the scene.

Hence, if the depth of the object changes enough, then the object is well removed.

More formally, we report the ratio of the object's pixels which depth changes by more than a threshold $\xi_{\text{depth}}$.
The threshold $\xi_{\text{depth}}$ is derived automatically with Generalized Histogram Thresholding~\cite{BarronECCV2020} on the histogram of depth differences over the whole image (See~\cref{fig:depth}).

The depth maps are derived from the scene's rendering before and after removal so they have consistent scales.
The defined ratio can be interpreted as the accuracy in depth change and is noted $\text{acc}_{\Delta\text{depth}}$:
\begin{equation}
    \text{acc}_{\Delta\text{depth}}^{\uparrow} = \frac{\# \text{object pixel with depth change} > \xi_{\text{depth}} }{\# \text{object pixels} }
\end{equation}
The object's pixel locations are specified by the object's mask in the image that we assume is available.

\subsection{Removal Refinement} 

Parts or residuals of an object can remain in the scene, even after removal.
To address this issue, we propose a graph-based refinement that leverages the tight coupling between the geometry and the semantics of the scene~\cite{mousavian2016joint,liu2019end,adam2023has}.

After a method outputs a set of Gaussians to be removed, the refinement looks for other Gaussian candidates for removal within the spatial and semantic neighborhood of that set.
The refinement defines a graph over all 3D Gaussians and partitions them into one set to be removed and one to remain.
The sets are initialized with the output of the removal method and refined to optimize the semantic and spatial consistency of the partitions.
This is done by finding a graph cut with maximum spatial and semantic consistency~\cite{boykov2001fast,zabih2004spatially,greig1989exact}.
Such a graph-based optimization is also used in GaussianCut~\cite{jain2024gaussiancut} to segment the foreground and background elements in the scene.
The proposed refinement differs in the graph definition and the energy cost.
In~\cite{jain2024gaussiancut} the correlation between two Gaussians depends on their spatial proximity and their color similarity.
Here, the correlation between two Gaussians depends not only on their spatial proximity, as in~\cite{jain2024gaussiancut}, but also their semantic similarity.
Next, we formalize the graph definition and the energy cost minimized in the graph-cut algorithm.

The graph's nodes are the 3D Gaussians and edges are drawn between spatially close Gaussians.
To reduce the size of the graph optimization, we filter out the Gaussians that do not `intersect' with the initial set of Gaussians to be removed.
Two Gaussians are said to `intersect' if the Euclidean distance between their centers is smaller than the sum of their largest scale.
This is equivalent to approximating the Gaussians as spheres and testing for the spheres' intersections.
Next, the edges are based on the spatial distance and feature similarity between the nodes.
The spatial distance between two Gaussians is defined as the Euclidean distance between their centers.
A node has edges with the nodes among its $K$ nearest neighbors that also have a feature similarity higher than a threshold $\delta$. 
The partition that optimizes the spatial and semantic consistency between nodes is the partition that minimizes the energy cost defined with two terms: the unary terms represent the likelihood of a Gaussian to be removed, and the binary term represents the semantic consistency between neighboring Gaussians.
The unary term is initialized with respectively 1 or 0 based on whether the Gaussian is removed or not by the method being refined.
The binary term between two Gaussian is their semantic similarity measured with the L2 distance between their semantic features.

\begin{table*}[t]
\small
\centering
  \begin{tabular}{llccc|cccc|cccc}
    \toprule
       Scene-Object & \multicolumn{4}{c}{IoU$_{\text{drop}}\uparrow$} 
        & \multicolumn{4}{c}{acc$_{\Delta\text{depth}}\uparrow$}  
        & \multicolumn{4}{c}{sim$_{\text{SAM}}\downarrow$}  \\
        \cmidrule(lr){2-13}
        & FGS & GG & SAGS & GC 
         & FGS & GG & SAGS & GC
          & FGS & GG & SAGS & GC
          \\
    \cmidrule(lr){1-13}
    Counter-Baking Tray & 0.34 & \U{0.53} & 0.10 & \B{0.62} & 0.99 & 0.96 & 0.21 & 0.98 & \B{0.21} & \U{0.35} & 0.71 & \U{0.35} \\
    Counter-Plant       & 0.75 & \U{0.84} & 0.03 & \B{0.86} & 1.00 & 1.00 & 0.01 & 0.99 & 0.13& \B{0.12} & 0.85 & \U{0.13} \\
    Counter-Gloves      & 0.01 & \B{0.60} & 0.10 & \B{0.60} & 0.01 & 1.00 & 0.55 & 1.00 & 0.99 & \B{0.12} & 0.56 & \U{0.16} \\
    Counter-Egg Box     & 0.08 & \B{0.63} & 0.56 & \U{0.62} & 0.06 & 1.00 & 0.86 & 1.00 & 0.84 & \B{0.15} & 0.47 & \U{0.19} \\
    Room-Plant & \B{0.53} & 0.26 & 0.17 & \B{0.53} & 0.97 & 0.70 & 0.33 & 0.99 & \U{0.22} & 0.33 & 0.57 & \B{0.14} \\
    Room-Slippers    & 0.00 & \B{0.82} & 0.25 & \U{0.48} & 0.00 & 1.00 & 0.91 & 0.98 & 1.00 & \B{0.05} & 0.45 & \U{0.35} \\
    Room-Coffee table& 0.57 & \B{0.86}& 0.00 & \B{0.86} & 0.67 & 0.89 & 0.06 & 0.99 & 0.26 & \U{0.08} & 0.86 & \B{0.07} \\
    Kitchen-Truck       & 0.62 & 0.61 & \B{0.67} & \U{0.66} & 0.96 & 1.00 & 1.00 & 0.92 & 0.35 & \U{0.17} & 0.22 & \B{0.08} \\
    \cmidrule(lr){1-13}
    Garden-Table  & 0.67 & 0.48 & \U{0.81} & \B{0.86} & 0.99 & 1.00 & 0.98 & 1.00 & 0.11& 0.14 & \B{0.04}& \U{0.06}\\
    Garden-Ball        & 0.00 & 0.16 & \U{0.41} & \B{0.42} & 0.00 & 0.60 & 0.60 & 0.53 & 0.59& \B{0.01}  & \U{0.21}& 0.37\\
    Garden-Vase        & 0.79 & 0.64 & \U{0.96} & \B{0.97}& 0.99 & 1.00 & 0.96 & 1.00 &  0.12& \B{0.10} & \U{0.11}& \U{0.11}\\
    \bottomrule
  \end{tabular}
  \caption{\textbf{Object removal evaluation with the proposed metrics.}
  The three metrics measure changes in semantics and depth before and after removal: 
  the IoU$_{\text{drop}}^\uparrow$ measures the drop in semantic segmentation, acc$_{\Delta\text{depth}}^\uparrow$ the changes in the depth maps, and sim$_{\text{SAM}}^\downarrow$ a change in the SAM~\cite{kirillov2023segany} masks.
  The \B{best} and \U{second-best} are highlighted for {IoU$_{\text{drop}}^\uparrow$} and {sim$_{\text{SAM}}^\downarrow$}.
  GaussianCut (GC)~\cite{jain2024gaussiancut} and GaussianGrouping(GG)~\cite{gaussian_grouping} mostly outperform Feature3DGS(FGS)~\cite{zhou2024feature} and SAGS~\cite{hu2024semantic}.
  }
  \label{tab:sematic_iou_compact}
\end{table*}

\section{Experiments}
\label{sec:eval}

We compare various Gaussian-based scene representations under the proposed metrics on indoor and outdoor scenes.

\PAR{Approaches.} 
We evaluate the \B{Feature3DGS (FGS)}~\cite{zhou2024feature} variant that distills the LSEG~\cite{lilanguage} semantic features aligned with CLIP's text features~\cite{radford2021learning}.
FGS is prompted with a tuple of text entries: one positive query is associated with the object of interest and the others are negative queries.
The search compares the Gaussians' feature with the features of each text entry, and their similarity is normalized.
The Gaussian is removed if the similarity with the query text prompt is higher than $0.4$.

\B{GaussianGrouping (GG)}~\cite{gaussian_grouping} distills SAM~\cite{kirillov2023segany} semantic features that operate at a finer granularity than LSEG~\cite{li2024segment}.
Also, GG enforces spatial consistency between semantically similar Gaussians so that close-by Gaussians have similar features.
Prompting is less intuitive than in FGS~\cite{zhou2024feature}, as one can only query one of the SAM instance labels present in the training views.
A Gaussian is removed if its feature is associated with the prompted instance label.
Post-processing then removes all Gaussians within the convex hull of the removed Gaussians.

\B{SAGS}~\cite{hu2024semantic} is a training-free and feature-free method that takes object masks as prompts.

It estimates a removal probability for each Gaussian based on projective geometry.

The 3D center of the Gaussian is projected on the images and the removal probability is the ratio of images in which the projections land on the object's location.

If the probability is higher than $0.7$, the Gaussian is removed.
SAGS regularizes the removal to better fit the actual object's boundaries by splitting the Gaussians around the boundary and only removing the splits within the boundary.
Although efficient, using projective geometry instead of splatting~\cite{kerbl3Dgaussians} when assigning Gaussians to the object does not account for the Gaussian's opacity, which may lead to over-removal.

\B{GaussianCut (GC)}~\cite{jain2024gaussiancut} is also a feature-free and training-free method that leverages the spatial and color correlations between Gaussians.

Given a trained 3DGS~\cite{kerbl3Dgaussians} scene, it models the scene as a graph and determines which Gaussians should be removed via graph optimization.

As for SAGS, the prompt is a set of object masks.

The Gaussians define the nodes of the graph and are extended with a single parameter representing the probability of the Gaussian to be removed.
The parameter is initialized by lifting the 2D prompt mask to 3D and refined via graph-cut optimization where the unary term represents the likelihood of the Gaussian to be removed and the binary term measures the color similarity and spatial distance between two Gaussians.

\PAR{Dataset.}
The methods are compared on indoor and outdoor scenes of the mipNeRF360 dataset~\cite{barron2022mip}.
The dataset provides images with their camera poses and calibration but no ground-truth semantic masks necessary to derive the proposed metrics.
Hence, we derive pseudo-ground-truth semantic masks for a set of objects using SAM~\cite{kirillov2023segany} and human annotations (see Supp.).

\PAR{Implementation Details.}
Some methods~\cite{gaussian_grouping,zhou2024feature} have large memory requirements that go beyond our computational resources. 
Thus, we select between 100 and 150 images from each scene.
This does not affect the trained representations that can be trained from as little as 60 images~\cite{jain2024gaussiancut} or 80 images~\cite{zhou2024feature}.
All methods are trained on the same set of images and the same input 3D point cloud derived with \gls{sfm} using COLMAP~\cite{schonberger2016structure}.
The graph optimization uses the pcp library~\cite{Raguet18,raguet2015preconditioning}.
The parameters of the graph refinement and of the evaluated methods are reported in the Supp.
We use GroundedSAM~\cite{kirillov2023segany,liu2023grounding,ren2024grounded} to derive the metrics based on semantic segmentation.

\begin{figure*}[t]
    \centering
    \includegraphics[width=\linewidth]{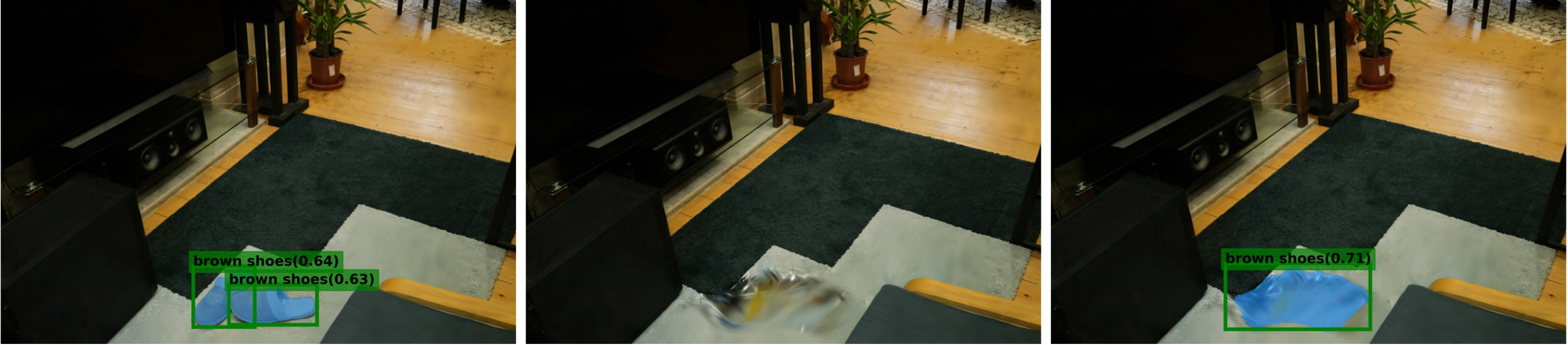}
    \caption{\textbf{Segmentation changes before and after removal.}
    Left-right: GroundedSAM~\cite{kirillov2023segany,liu2023grounding,ren2024grounded} overlay on the rendering before removal, rendering after removal, overlay after removal.
    The object is removed successfully but the segmentation model still finds it.
    One explanation can be that the pixel distribution on the edited area still exhibits patterns characteristic of the object, similar to what occurs in adversarial attacks.
    Even though the object can not be recognized by a human, GroundedSAM~\cite{kirillov2023segany,liu2023grounding,ren2024grounded} still manages to segment the object.
    }
    \label{fig:adversarial}
\end{figure*}

\subsection{Results}

\begin{table}[t]
\small
\centering
  \begin{tabular}{llccc} 
    \toprule
        & \multicolumn{4}{c}{acc$_{\text{IoU-post} > 0.5}^\downarrow$}  \\
        \cmidrule(lr){2-5}
        & FGS & GG & SAGS & GC 
          \\
    \cmidrule(lr){1-5}
    Baking Tray & 21.7 & \U{8.5} & 51.9 & \B{0.09} \\
    Plant       & \B{0}   & \B{0.0} & 83.3 & \B{0} \\
    Blue Gloves & 72.1 & \B{16.3} & 66.3 & \U{17.3} \\
    EggBox      & 80.4 & \B{0.0} & 4.1 & \U{1}     \\
    Plant (room)& \B{0} & 20 & 28 & \B{0}    \\
    Slippers    & 98.5 & \B{14.7} & 72.1 & \U{55.9} \\
    Table       & 38.4 & \B{1} & 90.9 & \B{1}     \\
    Truck       & 4.9 & 7.8 & \B{0}     & \U{0.05}  \\
    Table       & 29.7 & 45.9 & \U{12.2} & \B{4.7}  \\
    Ball        & 6.1 & \B{0} & \B{0} & \B{0}   \\
    Vase        & 10.8 & 21.6 & \B{0} & \B{0}  \\
    \bottomrule
  \end{tabular}
  \caption{\textbf{Object removal evaluation} based on the segmentation after removal only.
  We report the ratio acc$_{\text{IoU-post} > 0.5}\downarrow$ of images in which the segmentation still segments the object with IoU $>$ 0.5.
    A low value signals that the object is not found but says nothing about the reliability of the segmentation contrary to the proposed IoU$_{\text{drop}}^\uparrow$.
  }
  \label{tab:sematic_iou_tresh}
\end{table}

\PAR{Methods comparison.}
\cref{tab:sematic_iou_compact} evaluates the considered approaches under our proposed metrics.
The methods' ranks remain stable across the three metrics: 
GaussianCut (GC)~\cite{jain2024gaussiancut} and  GaussianGrouping (GG)~\cite{gaussian_grouping} achieve the best performances followed by SAGS~\cite{hu2024semantic} and Feature3DGS (FGS)~\cite{zhou2024feature}.
GG and GC are mostly on par with GC reaching slightly higher IoU$_{\text{drop}}^\uparrow$ whereas GG has better sim$_{\text{SAM}}^{\downarrow}$.
SAGS exhibit variance in the results: it gets on par with GC and even outperforms GG on the outdoor Garden scene but is often subpar on the rest of the scenes and objects.
The performance of FGS also varies, although it does not appear correlated to the type of scene.

One property that GG~\cite{gaussian_grouping} and GC~\cite{jain2024gaussiancut} share is leveraging the spatial correlation of Gaussians to assess whether they belong to the same object.
Their strong performance suggests that spatial consistency is an important property for scenes to be suitable for search and edit operations.
One advantage of GaussianCut~\cite{jain2024gaussiancut} over GG~\cite{gaussian_grouping} is its simplicity, and its memory and computational efficiency: it relies on a simple graph-cut optimization that can run on CPU and does not require distilling and storing semantic features into the 3DGS.

One explanation for the underperformance of FGS~\cite{zhou2024feature} may be that it is sensitive to the prompt.
For example, if the tuple of text prompts is not contrastive enough or if the prompts are out of distribution, the similarity between the Gaussian's feature and the prompt could be meaningless.

Another noticeable result is the gap between the indoor and outdoor performances of SAGS~\cite{hu2024semantic}.
One explanation may be that the objects outdoors are well-separated spatially so the projective-geometry-based association between objects and Gaussians is accurate enough.

\PAR{Complementarity analysis of the metrics.}
The ranking of the methods between the different metrics is mostly consistent, which is expected since they are all designed to measure the removal quality (\cref{tab:sematic_iou_compact}).
The presence of redundancy in the metrics makes the proposed evaluation robust to potential errors in the semantic models used in the derivation.
When a method achieves good results on all three metrics, then it is very likely that the removal succeeded and that the metrics are reliable.
However, a mix of good and bad metric scores indicates that either the removal quality is low or that the segmentation used to derive IOU$_\text{drop}^{\uparrow}$ and {sim$_{\text{SAM}}^\downarrow$} are incorrect, thus, the metric is not reliable. 

For example, a low IOU$_\text{drop}^{\uparrow}$ alone can be interpreted in two ways: the removal failed or the segmentation model is not reliable.
The latter happens when the object was not segmented even before the removal, \ie, IOU$_\text{pre}$ is low., and IOU$_\text{drop}^{\uparrow}$ is bound by this value, as discussed in~\cref{sec:method_iou_drop}.
The other metrics help disambiguate between the two interpretations.
In particular, the acc$_{\Delta\text{depth}}^{\uparrow}$ is a trustworthy indicator for failure cases because it tends to overestimate the quality of the removal.
Then, when this metric is low, it is likely that the removal failed, \eg, the performance of SAGS~\cite{hu2024semantic} on the `Baking Tray', `Plant' or `Coffee Table'.

Inversely, a high acc$_{\Delta\text{depth}}^{\uparrow}$ does not imply that the object is well removed since it overestimates the removal quality.

This is the case for GaussianGrouping~\cite{gaussian_grouping} on the `Garden-ball': the IOU$_\text{drop}^{\uparrow}$ at 0.16 is low, whereas the acc$_{\Delta\text{depth}}^{\uparrow}$ is at 0.60.
The third metric can disambiguate such a case: in this example, the sim$_{\text{SAM}}^{\downarrow}$ at 0.01 suggests that the removal performs well.
Then the low IOU$_\text{drop}^{\uparrow}$ does not indicate a low-quality removal but instead that the semantic segmentation is not reliable on this instance, which we visually confirmed.

This example illustrates how complementary the metrics are: they make the evaluation robust by catching the other metrics' failure cases.

We compare the proposed metrics to a more intuitive metric: the ratio of images in which the object is still segmented out after removal, and the lower the ratio, the better the removal.
An object is said to be segmented if the IoU is higher than a threshold in $\{0.5, 0.7, 0.9\}$.
We report the metric as acc$_{\text{IoU-post} > 0.5}^{\downarrow}$ for $0.5$ in~\cref{{tab:sematic_iou_tresh}} and for the other thresholds in the supplementary.
The results show that the acc$_{\text{IoU-post} > 0.5}^{\downarrow}$ is correlated to the IOU$_\text{drop}$, which supports that the latter is also a measure of the removal quality.
However, the IOU$_\text{drop}$ is more informative since it not only reports whether an object is removed but also whether the metric itself is reliable.

For example, FGS~\cite{zhou2024feature} achieves an acc$_{\text{IoU-post} > 0.5}^{\downarrow}$ of 6.1 on the `Garden-ball', \ie, the object is not segmented out after removal.
This could mean that the object removal is successful.
However, the IoU before removal is also low, \ie, the segmentation model can not find the object even in the original scene.
In that case, it is the segmentation that is not reliable, which is signaled with a low IOU$_\text{drop}^{\uparrow}$ of $0.0$: it tells that no conclusion can be drawn and the other metrics must be further analyzed.

\begin{table}[t]
\scriptsize
\centering
  \begin{tabular}{lcc|cc|cc}
    \toprule
      & \multicolumn{2}{c}{IoU$_{\text{drop}}\uparrow$} 
    & \multicolumn{2}{c}{acc$_{\Delta\text{depth}}\uparrow$}  
    & \multicolumn{2}{c}{sim$_{\text{SAM}}\downarrow$}  \\
    \cmidrule(lr){2-7}
    &  FGS & SAGS  &  FGS & SAGS  &  FGS & SAGS \\
        \midrule
    Counter-Plant           & +0    & *     & +0    & *     & +0    & *\\
    Counter-Egg Box         & *     & +0.04 & *     & +0.12 & *     & +0 \\ 
    Room-Plant              & +0.03 & +0.09 & +0.01 & +0.04 & -0.06 & +0 \\ 
    Room-Slippers           & *     & +0.17 & *     & +0    & *     & +0 \\ 
    Room-Coffee Table       & +0    & *     & +0.11 & *     & -0.14 & *\\
    Kitchen-Truck           & +0.04 & *     & +0    & *     & -0.21 & *\\
    Garden-Table            & *     & +0.09 & *     & +0.01 & *     & +0 \\ 
    Garden-Vase             & +0.11 & +0.01 & +0.01 & +0.04 & +0    & +0 \\ 
    \bottomrule
  \end{tabular}
  \caption{\textbf{Graph-based Removal Refinement.} We report the change in performance on FGS~\cite{zhou2024feature} and SAGS~\cite{hu2024semantic}.
  '+' indicates an improvement and '-' a drop in performance.
  Fail cases are reported with $*$ and occur when the refinement graph can not be built from the output of the removal method.
  }
  \label{tab:graphcut_results}
\end{table}

\PAR{Graph-based removal refinement.}
We evaluate the removal before and after refinement on the two methods that do not leverage spatial regularization: FGS~\cite{zhou2024feature} and SAGS~\cite{hu2024semantic}.
The results in~\cref{tab:graphcut_results} show that the refinement either improves or preserves the removal's quality measured by the IOU$_\text{drop}^{\uparrow}$ and the {acc$_{\Delta\text{depth}}^\uparrow$} but can hinder the sim$_{\text{SAM}}^\downarrow$.

Also, the refinement fails when the graph can not be initialized.
This occurs when the removal method being refined does not output a meaningful set of Gaussians to start from,\eg, FGS~\cite{zhou2024feature} does not remove the `Counter-eggbox' or the `Room-slippers' as indicated by their low IOU$_\text{drop}^{\uparrow}$ at $0.08$ and $0.0$.
The same occurs for SAGS~\cite{hu2024semantic} on the `Counter-plant' and the `Room-coffee table'.
The refinement can also fail outside of this scenario, \eg, FGS~\cite{zhou2024feature} on the `Garden-table'.
Still, the refinement remains relevant as one can decide to include it or not based on whether it improves the metrics.

\PAR{Qualitative Results.}
We show renderings of the evaluated methods before and after removal in~\cref{fig:adversarial,fig:sam,fig:depth}.
\cref{fig:adversarial} illustrates an interesting use-case where the object is not visible to humans anymore, yet GroundedSAM~\cite{kirillov2023segany,liu2023grounding,ren2024grounded} finds the object.
This suggests that barely visible information about the object can remain in the scene, even when the removal is successful to the human eye, and that the proposed metrics can detect such scenarios.
This opens interesting future directions on whether a network could be trained to invert the object removal from invisible pixel information and how to prevent it.

\begin{figure}
    \centering
    \includegraphics[width=\linewidth]{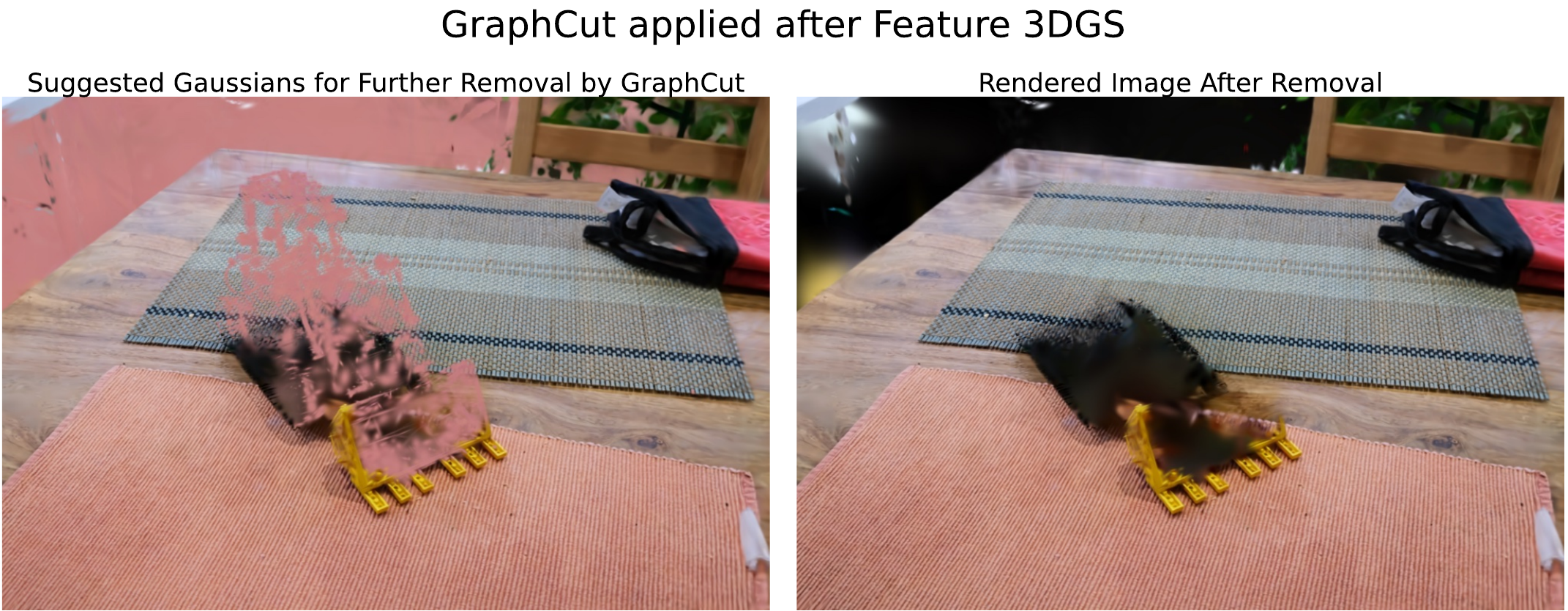}
     \includegraphics[width=\linewidth]{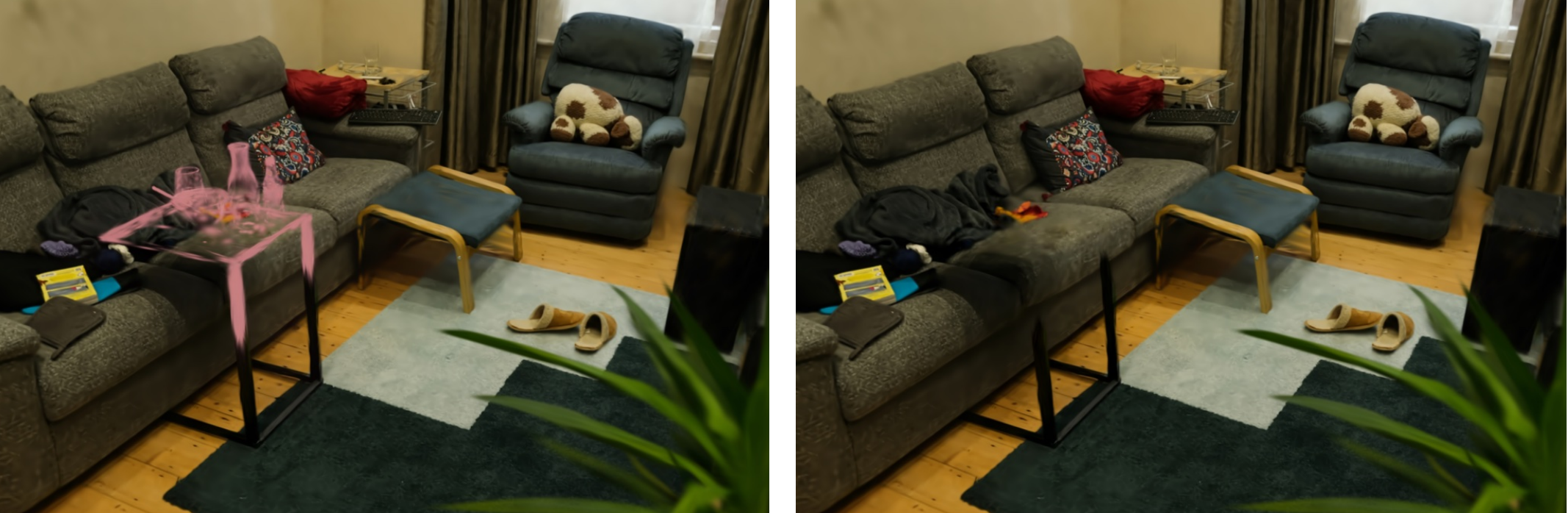}
    \caption{\textbf{Removal before and after refinement.} Left: removal before refinement. The area to be removed by the refinement is highlighted in pink. Right: removal after refinement.
    The refinement captures the object's boundaries (top) and even relevant affordances, such as the elements on the tables to be removed.
    }
    \label{fig:graph_cut}
\end{figure}

\cref{fig:graph_cut} shows the removal results of FGS~\cite{zhou2024feature} before (left) and after (right) the proposed graph-based refinement.
The Gaussians removed by the refinement are colored pink on the left image and remain within the boundaries of the object.
\cref{fig:sam} shows the distribution of SAM~\cite{kirillov2023segany} masks on the renderings of all methods before and after removal.
It provides a visual intuition on how sim$_{\text{SAM}}^{\downarrow}$ behaves: GG~\cite{gaussian_grouping} and GC~\cite{jain2024gaussiancut} remove the `plant' and the masks after removal and SAM~\cite{kirillov2023segany} now segments the objects behind the plant.
Conversely, the masks of FGS~\cite{zhou2024feature} and SAGS~\cite{hu2024semantic} before and after removal are similar because the removal failed.
\cref{fig:depth} provides a visual intuition for  acc$_{\Delta\text{depth}}^{\uparrow}$ with a success (top) and failure case (bottom): when the object is well removed, there is a change in the rendered depth before and after removal at the location of the removed object (outlined in green).
More visualizations are available in the supplementary.

\PAR{User Study.}
We complete the previous analysis with a user study with 8 volunteers. 
They evaluated the quality of the removal based on their ability to recognize the removed object, and their ability to detect whether an object has been removed.
The users are presented with renders before and / or after removal and must answer the following questions: 
(1) Was an object removed from this image?;
(2) Was an object removed from this specific area?;
(3) Which method best removes the object?;
(4) Does the graph-based refinement improve the removal?
The questions are worded in a way to reduce potential 
bias, \eg, 
in (1) and (2), the users are randomly shown images with no removal (see supplementary).

Overall, the conclusions of the study are consistent with the previous quantitative analysis: users find that GaussianCut~\cite{jain2024gaussiancut} performs best on 59\% of the images, followed by GG~\cite{gaussian_grouping} and SAGS~\cite{hu2024semantic} with 17\%, then FGS~\cite{zhou2024feature} with 7\%.
This is in line with the quantitative results reported in~\cref{tab:sematic_iou_compact}.
They also report that the graph-based refinement improves the removal.
The participants can detect whether an object is removed in $\sim$ 88\% of the images, irrelevantly of the removal method and whether the removal area is highlighted or not.
In the context of this paper, which focuses on the privacy aspect of the removal, a user being aware of the removal is not an issue as long as the removed content can not be identified.
This is in line with existing privacy regulations, \eg, the presence of blur in the images is fine as long as the faces and license plates are blurred~\cite{illman2019california,voigt2017eu}.

\PAR{Limitations.}
The metrics rely on off-the-shelf semantic segmentation models that are prone to errors.
Although introducing redundancy between the metrics alleviates this issue, it does not fully address it, which calls for further research on robust metrics for the evaluation of object removal.

\section{Conclusion}

We evaluate how well removal operations in 3DGS scenes perform in the context of privacy-preserving mapping by measuring whether there are residuals of the removed objects that can be reasoned over.
To do so, we introduce an evaluation framework that measures whether the object can still be recognized after removal using off-the-shelf semantic models and 3D perception with depth.
We test the proposed metrics on state-of-the-art scene representations, and results show that the proposed metrics are meaningful and consistent with a user study.

We also contribute a removal refinement based on spatial and semantic regularization.
We hope this evaluation will contribute to fostering research towards better privacy-preserving representations.

\section*{Acknowledgements}

This work was supported by 
the Czech Science Foundation (GACR) EXPRO (grant no. 23-07973X),
the Ministry of Education, Youth and Sports of the Czech Republic through the e-INFRA CZ (ID:90254),
the MIS 5154714 project of the National Recovery and Resilience Plan Greece 2.0 funded by the European Union under the NextGenerationEU Program.

\maketitlesupplementary
\appendix

The supplementary material provides additional implementation details, evaluation metrics, and extended experimental results supporting the main paper. 

Section A describes the implementation details. 
Subsection A.1 presents the parameter settings for the compared methods: Feature3DGS~\cite{zhou2024feature}, SAGS~\cite{hu2024semantic}, GaussianGrouping~\cite{gaussian_grouping}, and GaussianCut~\cite{jain2024gaussiancut}. 
Subsection A.2 details the evaluation metrics, including the generation of pseudo-ground-truth semantic masks.
Subsection A.3 details the graph-cut refinement process, including its formulation and parameter settings.

Section B presents additional qualitative and quantitative results. 
Subsection B.1 reports the user study that evaluates the object removal quality and the impact of the graph-based refinement. 
Subsection B.2 reports the evaluation results of graph-based refinement. 
Subsection B.3 reports detailed results from the semantic IOU evaluation and the depth difference after object removal.

\section{Implementation Details}

\subsection{Compared Methods}

We use the official implementations provided by the respective authors and apply the following settings.

\PAR{Feature3DGS (FGS)~\cite{zhou2024feature}.}
FGS is prompted with a tuple of text entries: one positive text prompt is associated with the object of interest and the others are negative text prompts.
The search compares the Gaussians' feature with the features of each text entry, and their similarity is normalized with softmax.
We use the following negative prompts per scene:
Garden: \{grass, sidewalk, tree, house\}, 
Room: \{sofa, rug, television, floor\}, 
Kitchen: \{rug, table, chair\},
Counter: \{oranges, wooden rolling pin, coconut oil\}.
A Gaussian is removed if the similarity between the Gaussian's feature and the prompt feature is higher than a threshold.
We set this similarity threshold to $0.4$ for all scenes and objects.

\PAR{SAGS~\cite{hu2024semantic}}
is a training-free and feature-free method that takes object masks as prompts.
It estimates a removal likelihood for each Gaussian based on projective geometry.
The 3D center of the Gaussian is projected on the images and the removal probability is the ratio of images in which the projections land on the object's location.
A Gaussian is removed if its likelihood to be removed is higher than $0.7$.

\PAR{GaussianGrouping~\cite{gaussian_grouping}.}
For the label training we set the SAM IoU prediction threshold to $0.8$.
For Gaussian training, we use default settings $\text{densify\_until\_iter}=10000$, $\text{num\_classes}=256$, $\text{reg3d\_interval}=5$, $\text{reg3d\_k}=5$, $\text{reg3d\_lambda\_val}=2$, $\text{reg3d\_max\_points}=200000$, $\text{reg3d\_sample\_size}=1000$.
For the object removal setting, we use the default number of classes of $256$ and the removal threshold of $0.3$. 

\PAR{GaussianCut~\cite{jain2024gaussiancut}}
is also a feature-free and training-free method that leverages the spatial and color correlations between Gaussians.
Given a trained 3DGS~\cite{kerbl3Dgaussians} scene, it models the scene as a graph and determines which Gaussians should be removed via graph optimization.
As for SAGS, the prompt is a set of object masks.
The Gaussians define the nodes of the graph and are extended with a single parameter representing the probability of the Gaussian to be removed.
The parameter is initialized by lifting the 2D prompt mask to 3D and refined via graph-cut optimization where the unary term represents the likelihood of the Gaussian to be removed and the binary term measures the color similarity and spatial distance between two Gaussians.
The graph is built with the following parameters: 
the number of edges is set to $10$, the terminal cluster source is set to $5$, the terminal cluster sink to $5$, and the leaf size to $40$. 
We use a foreground threshold of $0.9$ and the prompt is a set of multi-view masks associated with the object to be removed.

\subsection{Metrics}

\PAR{Pseudo-ground-truth semantic masks generation.}
MipNeRF360~\cite{barron2022mip} does not provide ground-truth semantic masks required for evaluation.
Thus, we generate pseudo-ground-truth masks by running SAM~\cite{kirillov2023segany} on images, segmenting all objects, and selecting the masks corresponding to the target objects. 
These masks are sufficiently accurate for evaluation. 
Since SAM~\cite{kirillov2023segany} sometimes produces incomplete segmentation, we exclude the images with incomplete segmentation and use only fully segmented images.

\PAR{Semantic Recognition.}
If Grounded-SAM~\cite{kirillov2023segany,liu2023grounding,ren2024grounded} fails to detect an object for a given prompt, no semantic mask is produced, and the semantic IoU is $0$.

\subsection{Graph-Cut Refinement}
We use FAISS~\cite{douze2024faiss,johnson2019billion} for nearest neighbor search in the graph refinement. 
The graph-cut algorithm~\cite{Raguet18,raguet2015preconditioning} minimizes a functional over a graph \( G = (V, E) \) with total variation (d1) penalization, a separable loss term (we use smoothed Kullback-Leibler divergence loss), and simplex constraints
\begin{equation}
F (x) = f (x) + \|x\|_{d1} + \iota_{\text{simplex}}(x),
\end{equation}
where:
\begin{itemize}
    \item \( x_v \in \mathbb{R}^D \) is the label for each vertex \( v \in V \),
    \item \( D = \{0,1\} \) represents labels for removal (d=1) or retention (d=0),
    \item \( f(x) \) is a separable data-fidelity loss,
    \item $\|x\|_{d1} = \sum_{(u,v) \in E} w_{d1_{uv}} \left( \sum_{d=0}^{D} w_{d1_{d}} |x_{ud} - x_{vd}| \right)$,
    \item $w_{d1_{uv}}$ is the weight of the edge between vertex $u$ and $v$,
    \item $w_{d1_{d}}$ is the diagonal metric of the weights belonging to $d$-label.
\end{itemize}
The simplex constraint is defined as:
\begin{equation}
\iota_{\text{simplex}}(x) =
\begin{cases} 
0, & \text{if } x_{vd} \geq 0 \text{ and } \sum_d x_{vd} = 1, \forall v, d \\  
\infty, & \text{otherwise}  
\end{cases}
\end{equation}

\PAR{Graph Construction and Parameters}
The graph nodes, edges, and cut parameters vary with respect to the scene, the removed object, and the removal method (Feature3DGS~\cite{zhou2024feature} or SAGS~\cite{hu2024semantic}).
The parameters used for the graph construction are reported in~\cref{tab:graphcut_parameter_setting}.

\begin{table}[t]
\scriptsize
\centering
  \begin{tabular}{lcc|cc}
    \toprule
      & \multicolumn{2}{c}{Nearest Neighbors $K$} 
    & \multicolumn{2}{c}{Feature Similarity Threshold $\delta$}\\
    \cmidrule(lr){2-5}
    &  FGS & SAGS  &  FGS & SAGS   \\
        \midrule
    Counter-Plant           & 10    & *     & 0.8   & *     \\
    Counter-Egg Box         & *     & 10    & *     & 0.8   \\
    Room-Plant              & 4     & 5     & 1.0   & 0.8   \\
    Room-Slippers           & *     & 4     & *     & 1.0    \\
    Room-Coffee Table       & 6     & *     & 1.0   & *     \\
    Kitchen-Truck           & 4     & *     & 1.0   & *     \\
    Garden-Table            & *     & 10    & *     & 0.8   \\
    Garden-Vase             & 4     & 10    & 1.0   & 0.8   \\
    \bottomrule
  \end{tabular}
  \caption{\textbf{Graph-based refinement parameters.}
  Graph parameters for different methods, scenes, and object removal. Failures, marked by “*”, occur when the constructed graph is empty, preventing further refinement.}
  \label{tab:graphcut_parameter_setting}
\end{table}

The cut-threshold for determining Gaussian removal (1) or retention (0) is automatically set at the 95th percentile of the removal probability distribution. 
This means that the top $5\%$ of Gaussian splats with the highest removal probabilities are removed, irrespective of their prior removal labels.
Additional Gaussians for removal are identified by comparing the graph-cut removal labels with those from Feature3DGS~\cite{zhou2024feature} and SAGS~\cite{hu2024semantic}, thereby refining the overall removal process.

\section{Additional Results}

\subsection{User Study}

\PAR{Motivation.}
We complete the previous analysis with a user study with 8 volunteers.
They evaluated the quality of the removal based on their ability to recognize the removed object, and their ability to detect whether an object has been removed.
The users are presented with renders before and / or after removal and must answer the following questions.
We describe the `Goal' of the question, \ie, what we evaluate, the `Display', \ie, what content is shown to the user, and the `Question' asked to the user.

\B{Q1.}
Goal: Assess the removal success and assess whether potential removal artifacts
are similar to ``regular'' artifacts in the renderings?
To avoid biasing the study by showing only images with removed elements, hence biasing the user to answer only `yes', images before removal are also shown with a random probability.

Display: an image of the render before or after object removal.

Question: Was an object removed from this image?

\B{Q2.}
Goal: Asses the removal success in a region of the image.

Display: an image of the render after object removal and with a box drawn around the object area so that the user focuses on that particular area.
Again, to avoid biasing the study by showing only images with removed elements, images before removal are also shown with a random probability.

Question: Was an object removed from this area?

\B{Q3.}
Goal: Compare the removal from several methods and assess whether the associated metrics on that image are consistent with the human assessment.

Display: an image of the render after object removal with a box drawn around the object area. 
We display the results from several methods next to each other.

Question: In which image is the object best removed?

\B{Q4.} 
Goal: Assess whether the proposed graph-cut refinement improves the removal.

Display: an image of the render after object removal but before refinement, and after both the removal and the refinement.

Question: In which image is the object best removed?

Results: 97\% of the users picked the image after graph-cut refinement, which supports that the refinement makes the removal visually better.

\begin{figure}
    \centering
    \includegraphics[width=0.7\linewidth]{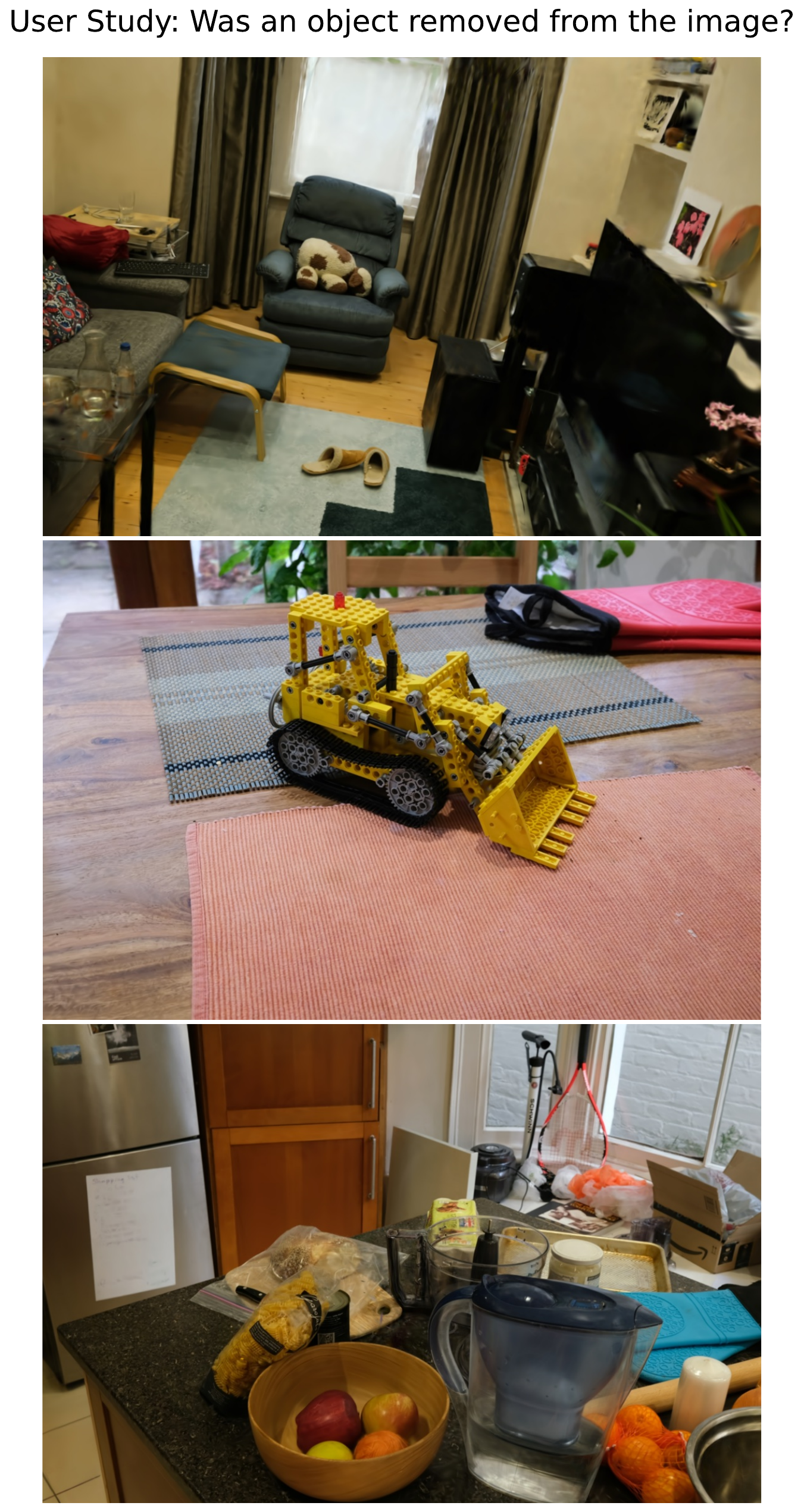}
    \caption{\textbf{User-study Q1: Was an object removed from this image?}.
    Examples of images shown to the study's users, top-bottom: yes is a correct answer (removal of the `Table'), no is a correct answer (no removal), and no is an incorrect answer (the `Plant' is removed but the removal is not detected by users).
    }
    \label{fig:q2}
\end{figure}

\begin{figure}
    \centering
    \includegraphics[width=0.7\linewidth]{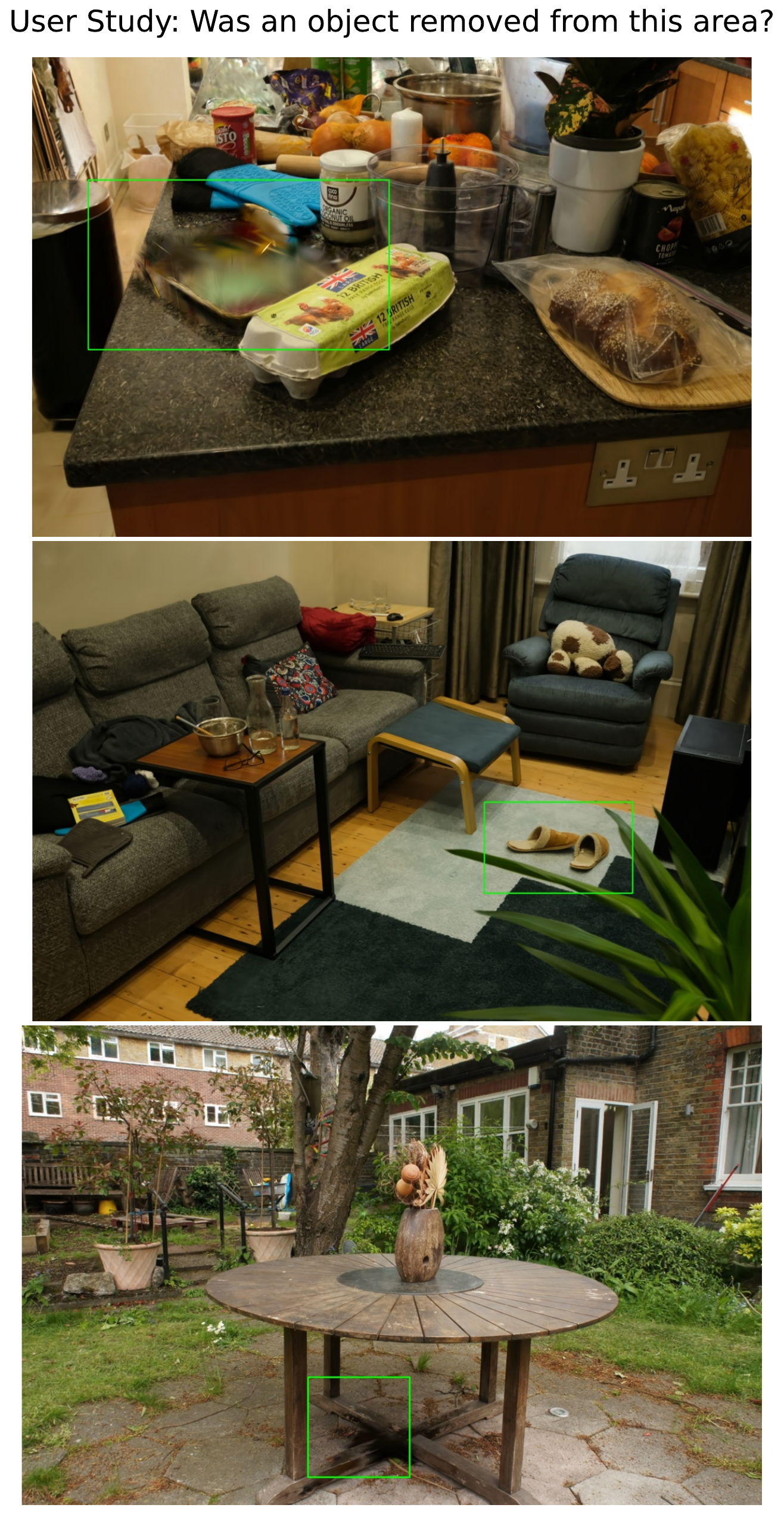}
    \caption{
    \textbf{User-study Q2: Was an object removed from this area?}.
    Examples of images shown to the study's users, top-bottom: yes is a correct answer (removal of the `Tray'), no is a correct answer (no removal), and no is an incorrect answer (the `Ball' is removed but the removal is not detected by the users).
    }
    \label{fig:q3}
\end{figure}

\begin{figure*}
    \centering
    \includegraphics[width=0.7\linewidth]{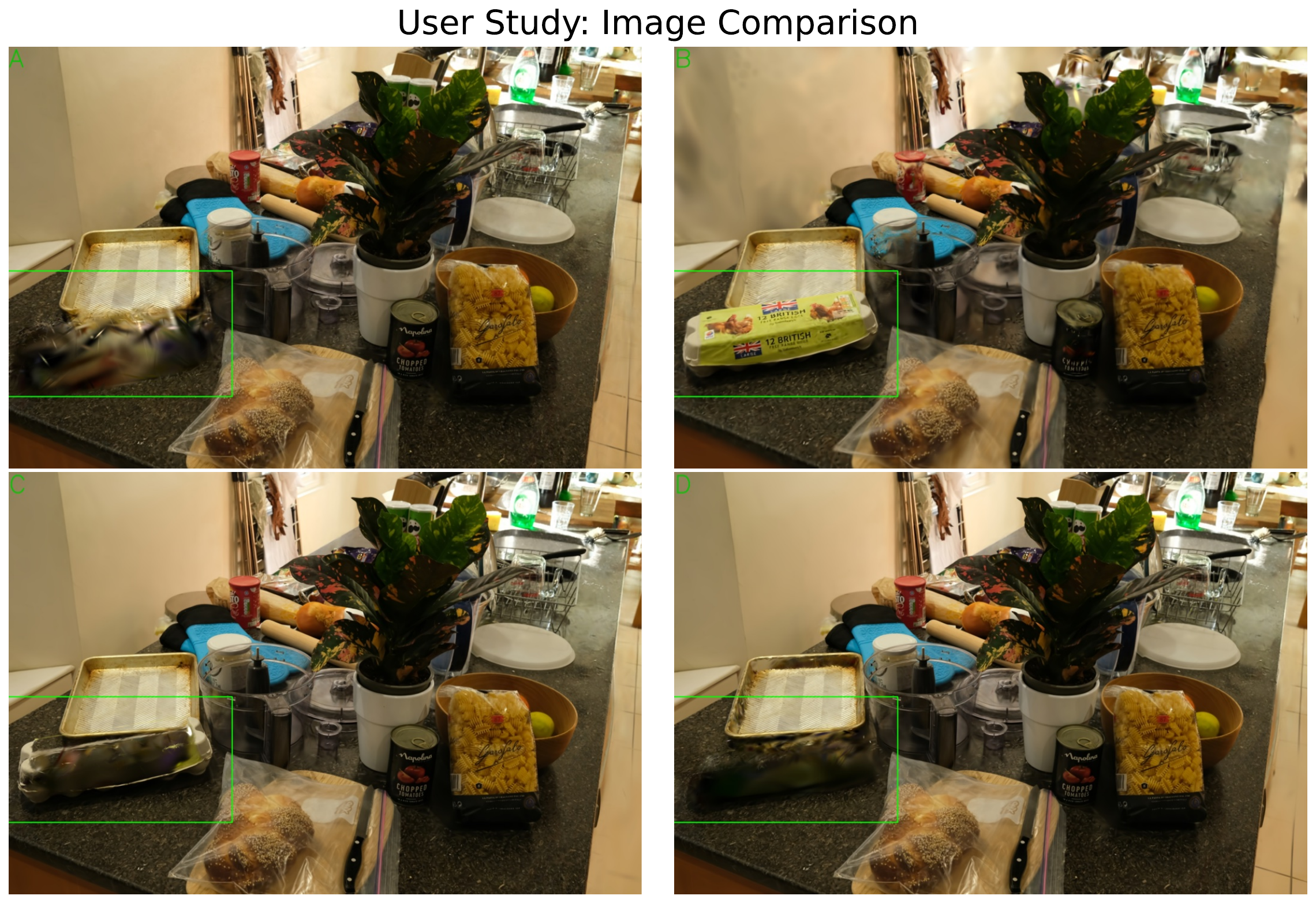}
    \caption{
    \textbf{User-study Q3: Comparison of different object removal methods.}
    The letters at the top left corner of each image are associated with one of the evaluated methods: GaussianGrouping~\cite{gaussian_grouping} (A), Feature3DGS~\cite{zhou2024feature} (B) (top row), SAGS~\cite{hu2024semantic} (C), and GaussianCut~\cite{jain2024gaussiancut} (D) (bottom row). 
    The images are renderings of the scene after removal.
    %
    Users are asked to pick the image with the most effective object removal, \ie, where the removed object is undetectable and the scene remains undisturbed.
    Despite similar quantitative results between GaussianGrouping~\cite{gaussian_grouping} (A) and GaussianCut~\cite{jain2024gaussiancut} (D), all users favored GaussianCut (D), demonstrating its superior visual performance in removing objects while preserving scene integrity.
    }
    \label{fig:q4_counter}
\end{figure*}

\begin{figure*}
    \centering
    \includegraphics[width=0.7\linewidth]{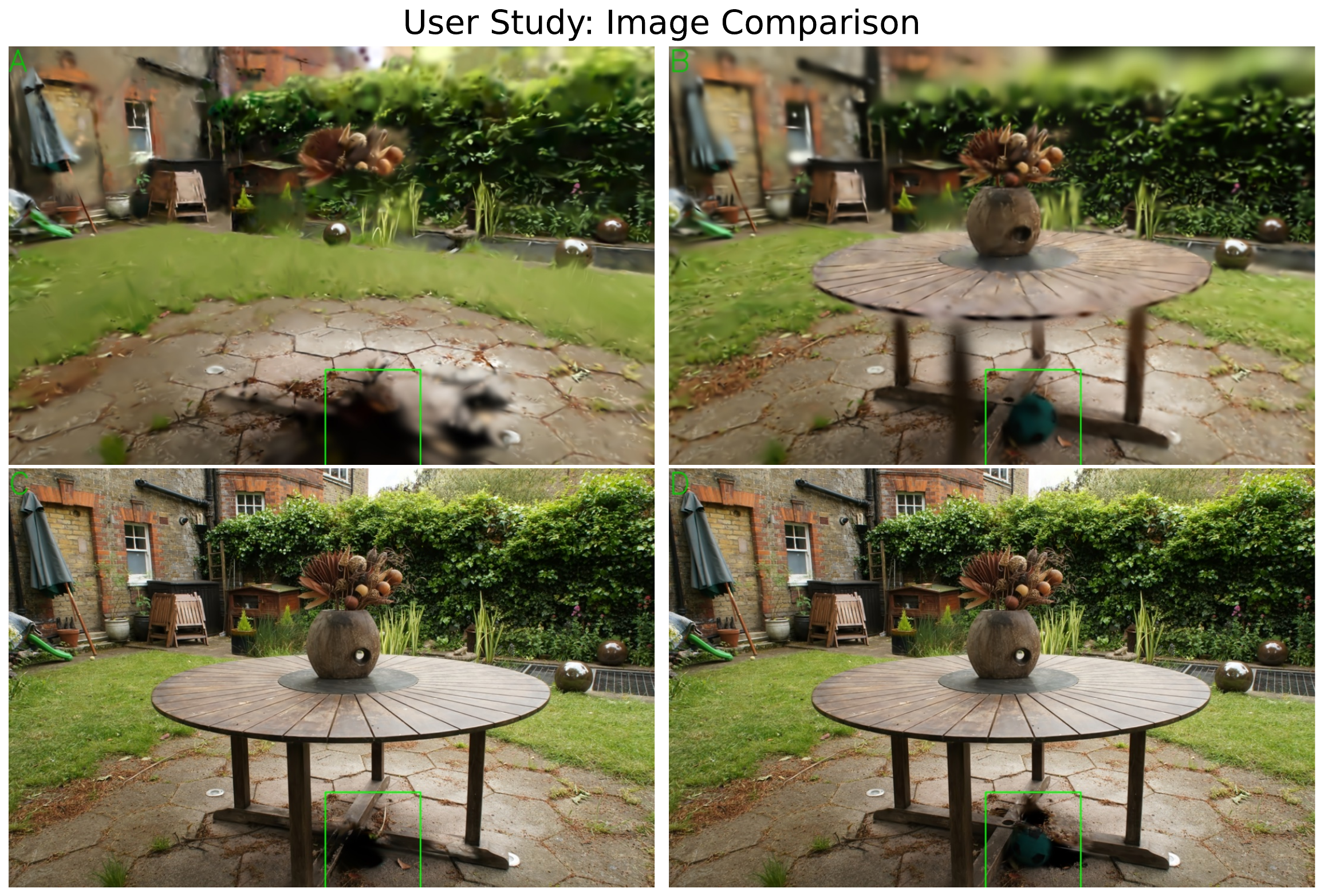}
    \caption{
    \textbf{User-study Q3: Comparison of different object removal methods.}
    The letters at the top left corner of each image are associated with one of the evaluated methods: GaussianGrouping~\cite{gaussian_grouping} (A), Feature3DGS~\cite{zhou2024feature} (B) (top row), SAGS~\cite{hu2024semantic} (C), and GaussianCut~\cite{jain2024gaussiancut} (D) (bottom row). 
    The images are renderings of the scene after removal.
    Users are asked to select the image with the most effective object removal, where the removed object is undetectable and the scene remains undisturbed. 
    Interestingly, all users favored SAGS (C), even though it did not achieve the best quantitative results, highlighting the importance of subjective evaluation in assessing visual quality.}
    \label{fig:q4_garden}
\end{figure*}

\begin{figure*}
    \centering
    \includegraphics[width=0.7\linewidth]{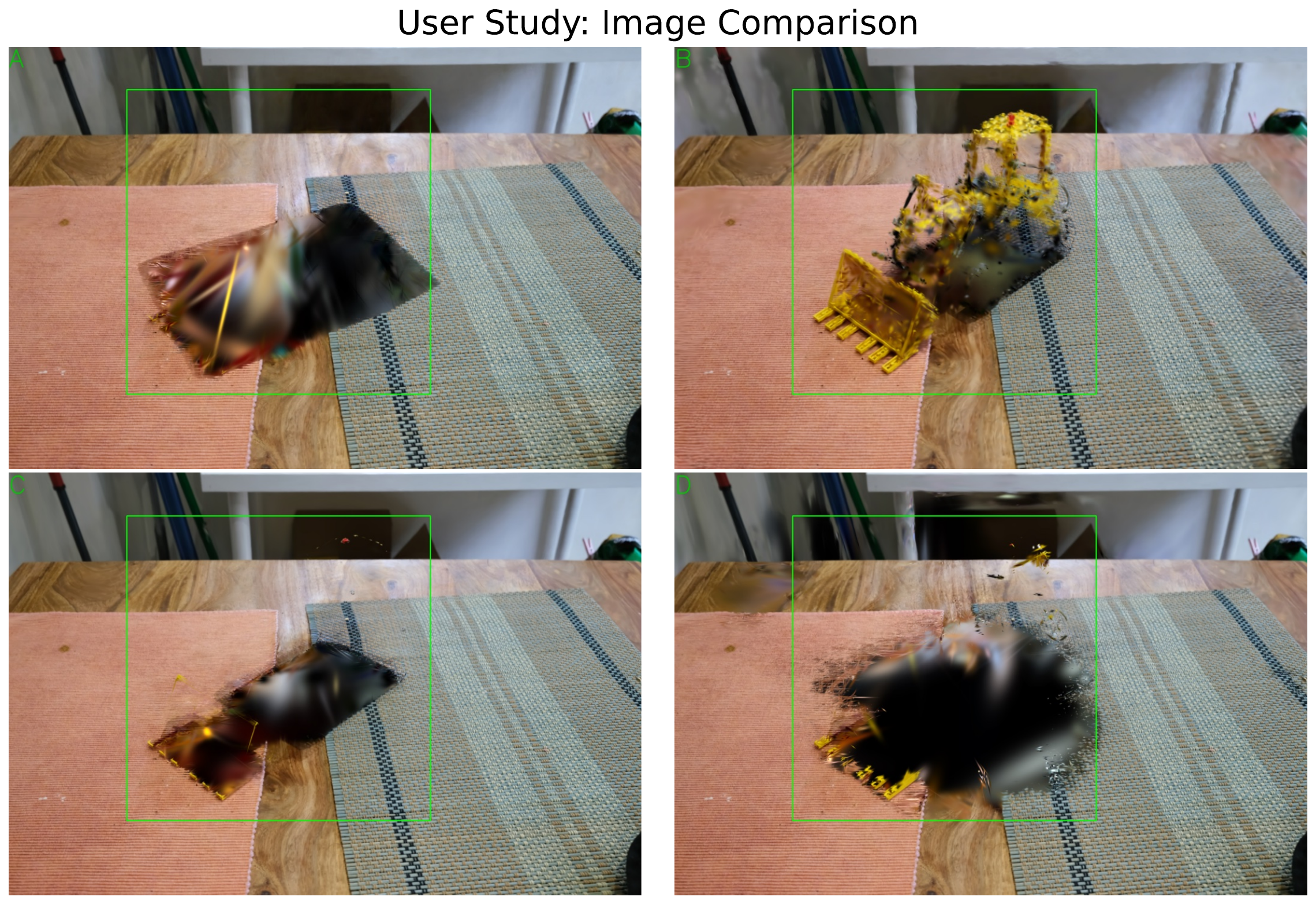}
    \caption{
    \textbf{User-study Q3: Comparison of different object removal methods.}
    The letters at the top left corner of each image are associated with one of the evaluated methods: GaussianGrouping~\cite{gaussian_grouping} (A), Feature3DGS~\cite{zhou2024feature} (B) (top row), SAGS~\cite{hu2024semantic} (C), and GaussianCut~\cite{jain2024gaussiancut} (D) (bottom row). 
    The images are renderings of the scene after removal.
    Users are asked to select the image with the most effective object removal, where the removed object is undetectable and the scene remains undisturbed. 
    Despite similar quantitative results between SAGS (C) and GaussianCut (D), all users favored SAGS (C), demonstrating its superior visual performance in seamlessly removing objects while preserving scene integrity.
    }
    \label{fig:q4_kitchen}
\end{figure*}

\begin{figure*}
    \centering
    \includegraphics[width=0.7\linewidth]{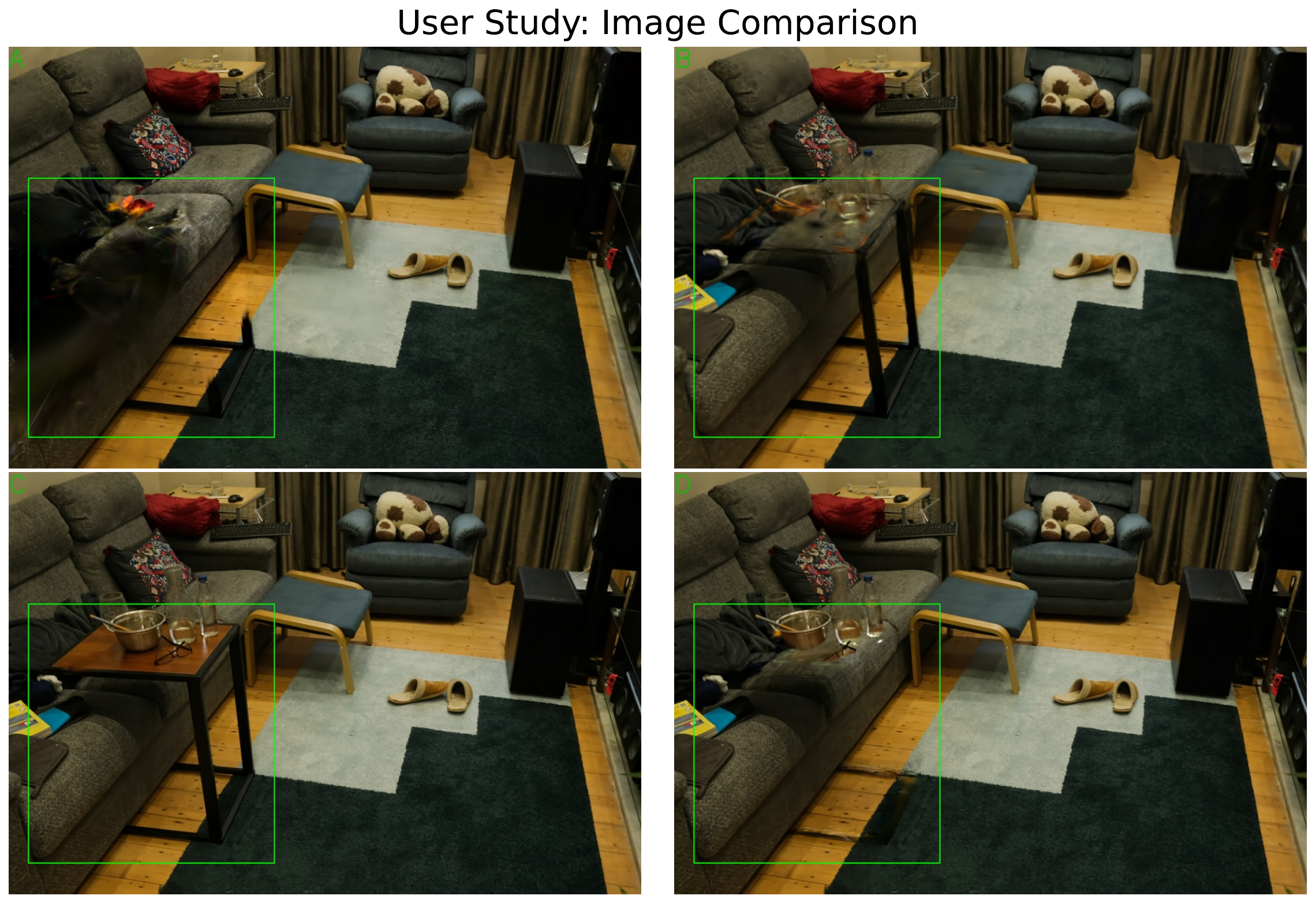}
    \caption{
    \textbf{User-study Q3: Comparison of different object removal methods.}
    The letters at the top left corner of each image are associated with one of the evaluated methods: GaussianGrouping~\cite{gaussian_grouping} (A), Feature3DGS~\cite{zhou2024feature} (B) (top row), SAGS~\cite{hu2024semantic} (C), and GaussianCut~\cite{jain2024gaussiancut} (D) (bottom row). 
    The images are renderings of the scene after removal.
    Users are asked to select the image with the most effective object removal, where the removed object is undetectable and the scene remains undisturbed. 
    Despite similar quantitative results between GaussianGrouping (A) and GaussianCut (D), all users favored GaussianCut (D), highlighting its superior visual performance in removing objects while preserving scene integrity.
    }
    \label{fig:q4_room}
\end{figure*}

\begin{figure*}
    \centering
    \includegraphics[height=0.8\textheight]{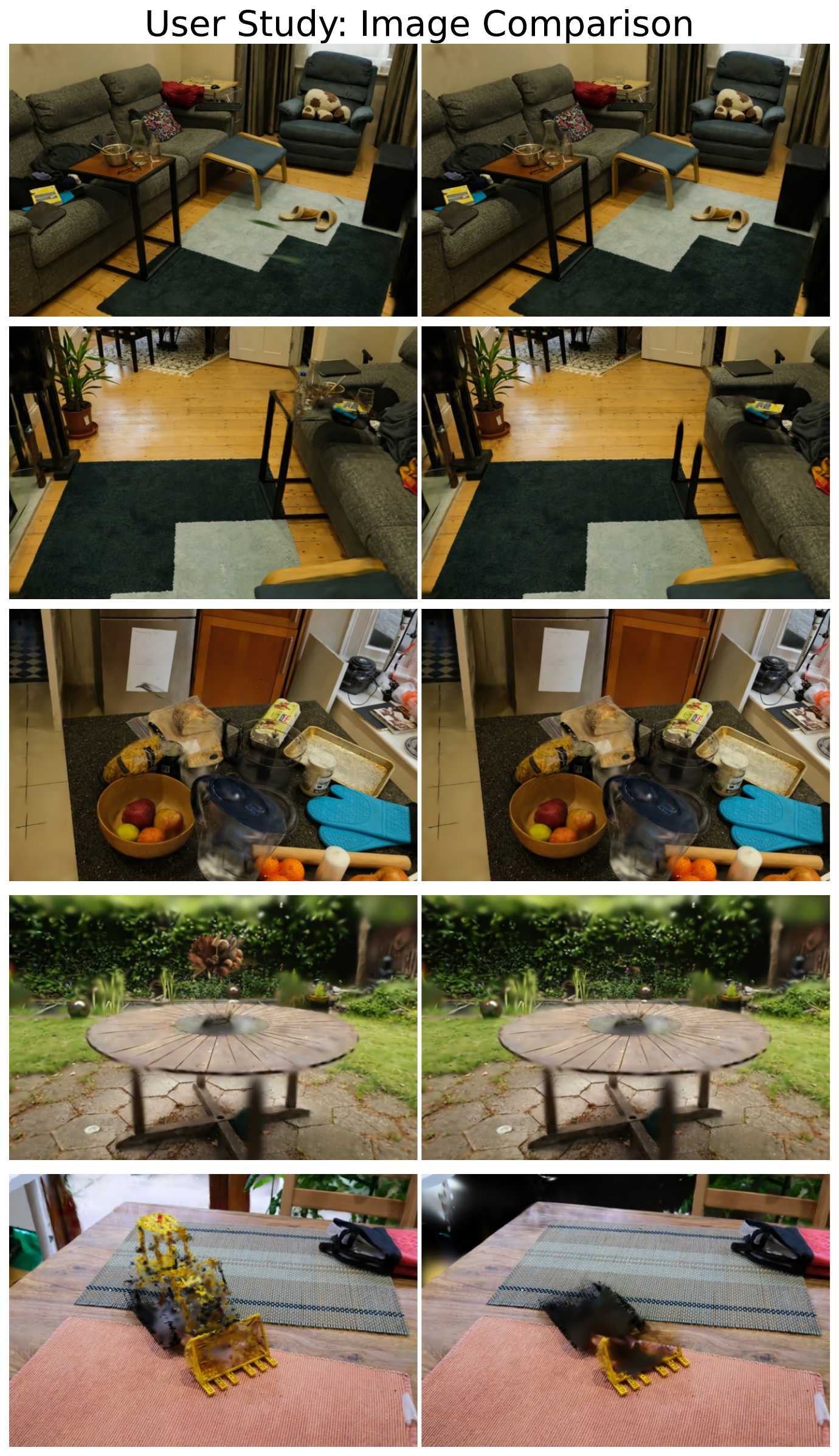}
    \caption{\textbf{User-study Q4: Graph-based refinement on Feature3DGS~\cite{zhou2024feature}.}
    Left-Right: before refinement, after refinement.
    All users prefer the refined removal, indicating that the proposed graph-based refinement improves the visual quality of object removal.}
    \label{fig:q5_feature3dgs}
\end{figure*}

\begin{figure*}
    \centering
    \includegraphics[height=0.8\textheight]{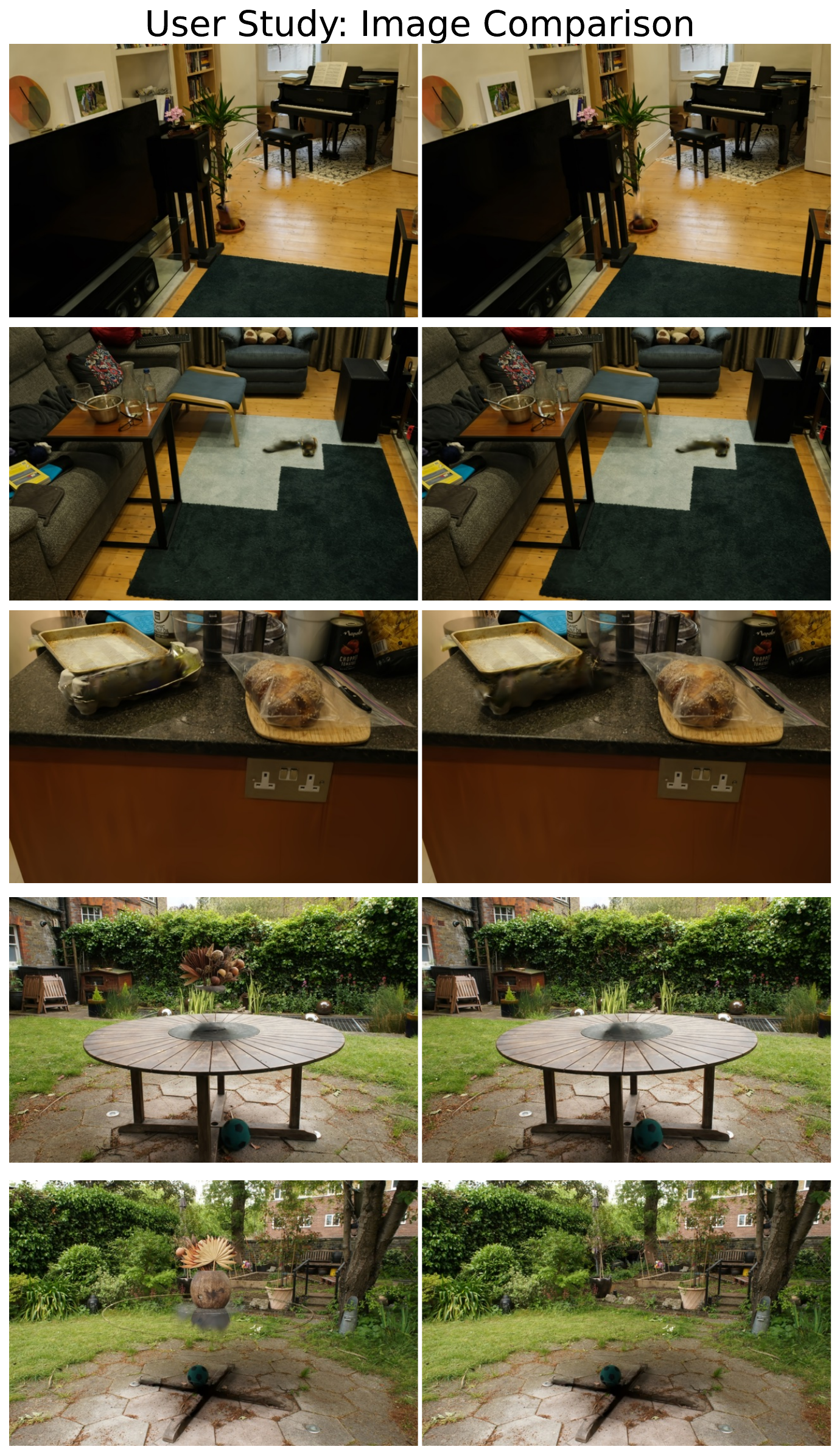}
    \caption{
    \textbf{User-study Q4: Graph-based refinement on SAGS~\cite{hu2024semantic}.} 
    Left-Right: before refinement, after refinement.
    All users prefer the refined removal, indicating that the proposed graph-based refinement improves the visual quality of object removal.}
    \label{fig:q5_sags}
\end{figure*}

\PAR{Results.} 
We show examples of images shown in the user study in~\cref{fig:q2} for Q1, in~\cref{fig:q3} for Q2, in~\cref{fig:q4_counter}-~\cref{fig:q4_room} for Q3, and in~\cref{fig:q5_feature3dgs}, ~\cref{fig:q5_sags} for Q4.
The caption describes the users' conclusion for each instance.
Overall, user responses align with the proposed metrics, which supports the soundness of the metrics.
\cref{tab:user_method_comparison} presents object-wise results aggregated from Q4: which method best removes the object? 
For each object, we report the ratio of votes accumulated over multiple images and all users.
`-' indicates that no user has voted for a given method.
The users often find that GaussianCut~\cite{jain2024gaussiancut} is the best followed by GaussianGrouping~\cite{gaussian_grouping} and SAGS~\cite{zhou2024feature}.

\begin{table}[t]
\centering
  \begin{tabular}{lcccc}
    \toprule
    &  FGS & GG & SAGS & GC \\
        \midrule
    Counter - Tray          & 0.14 & 0.05 & 0.05 & \B{0.76} \\
    Counter - Plant         & - & - & - & \B{1.0} \\
    Counter - Gloves        & - & 0.19 & - & \B{0.81} \\
    Counter - Egg Box       & - & - & - & \B{1.0} \\
    Room - Plant            & - & 0.24 & - & \B{0.76} \\ 
    Room - Slippers         & - & \B{0.67} & - & 0.33 \\
    Room - Coffee Table     & - & 0.24 & - & \B{0.76} \\
    Kitchen - Truck         & - & 0.29 & - & \B{0.71} \\
    Garden - Table          & \B{0.62} & 0.05 & - & 0.33 \\
    Garden - Ball           & - & - & \B{1.0} & - \\
    Garden - Vase           & - & 0.14 & 0.10 & \B{0.76} \\
    \bottomrule
  \end{tabular}
  \caption{\textbf{User-study Votes for the best removal methods} (normalized per object).
  Removal performance on various objects according to the user study (the higher, the better).
  The users are presented with renderings after object removal from various methods and they vote for the best method on each image.
  For each object, we report the ratio of votes accumulated over multiple images and all users.
  `-' indicates that no user has voted for a given method.
  The users often find that GaussianCut~\cite{jain2024gaussiancut} is the best followed by GaussianGrouping~\cite{gaussian_grouping} and SAGS~\cite{zhou2024feature}.
  }
  \label{tab:user_method_comparison}
\end{table}

\subsection{Graph-Based Refinement Results}
~\cref{tab:graphcut_results} reports the results of the removal with the proposed graph-based refinement.
The values between paratheses indicate the variation in performance where `+' and `-' signal an improvement and a deterioration respectively.
The evaluation runs on Feature3DGS~\cite{zhou2024feature} and SAGS~\cite{hu2024semantic}.

\begin{figure*}
    \centering
    \includegraphics[width=\linewidth]{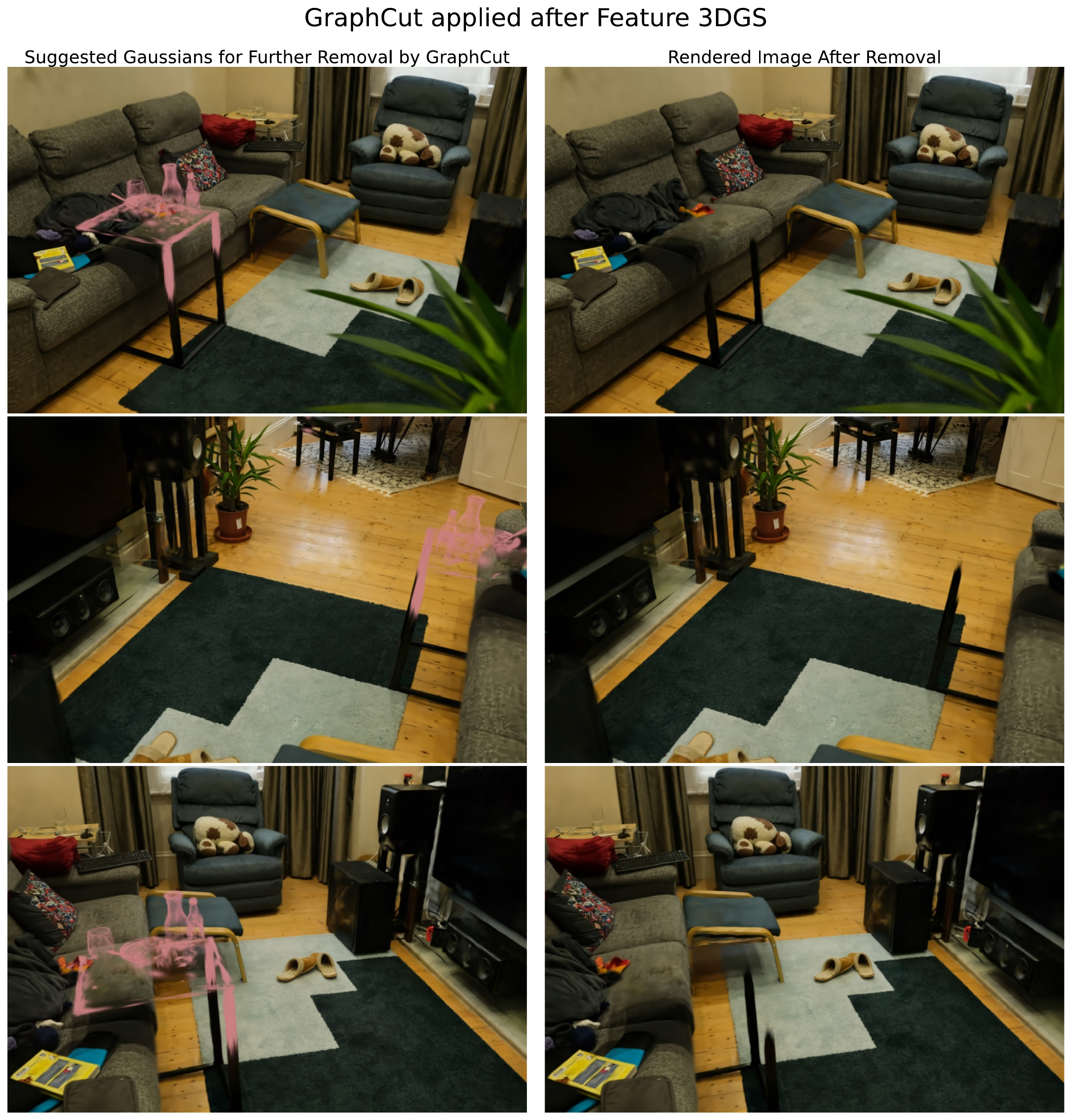}
    \caption{
    \textbf{Graph-based refinement of the removal on Feature3DGS~\cite{zhou2024feature}.}
    Left: the image shows the rendering after removal and before refinement.
    The Gaussians to be removed by the refinement are colored in pink.
    Right: the images are rendered after the refined removal.
    The improved removal of the `Table' in the right image demonstrates the effectiveness of the proposed graph-based refinement in identifying and eliminating residual parts left by the initial method.
    }
    \label{fig:graphcut_feature3dgs_room_table}
\end{figure*}

\begin{figure*}
    \centering
    \includegraphics[width=\linewidth]{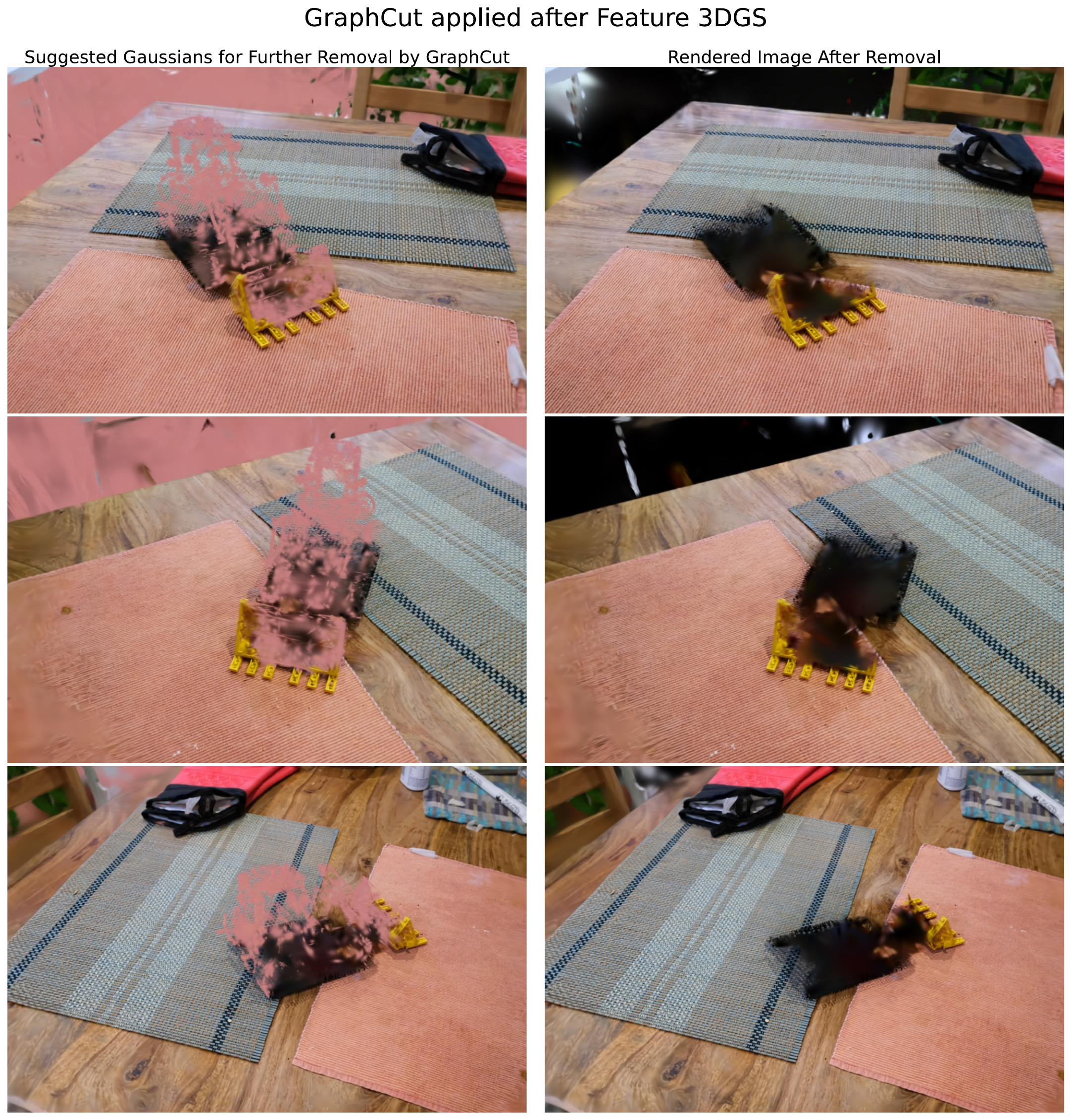}
    \caption{
    \textbf{Graph-based refinement of the removal on Feature3DGS~\cite{zhou2024feature}.}
    Left: the image shows the rendering after removal and before refinement.
    The Gaussians to be removed by the refinement are colored in pink.
    Right: the images are rendered after the refined removal.
    The improved removal of the `Truck' in the right image demonstrates the effectiveness of the proposed graph-based refinement in identifying and eliminating residual parts left by the initial method.
    }
    \label{fig:graphcut_feature3dgs_kitchen_truck}
\end{figure*}

\begin{figure*}
    \centering
    \includegraphics[width=\linewidth]{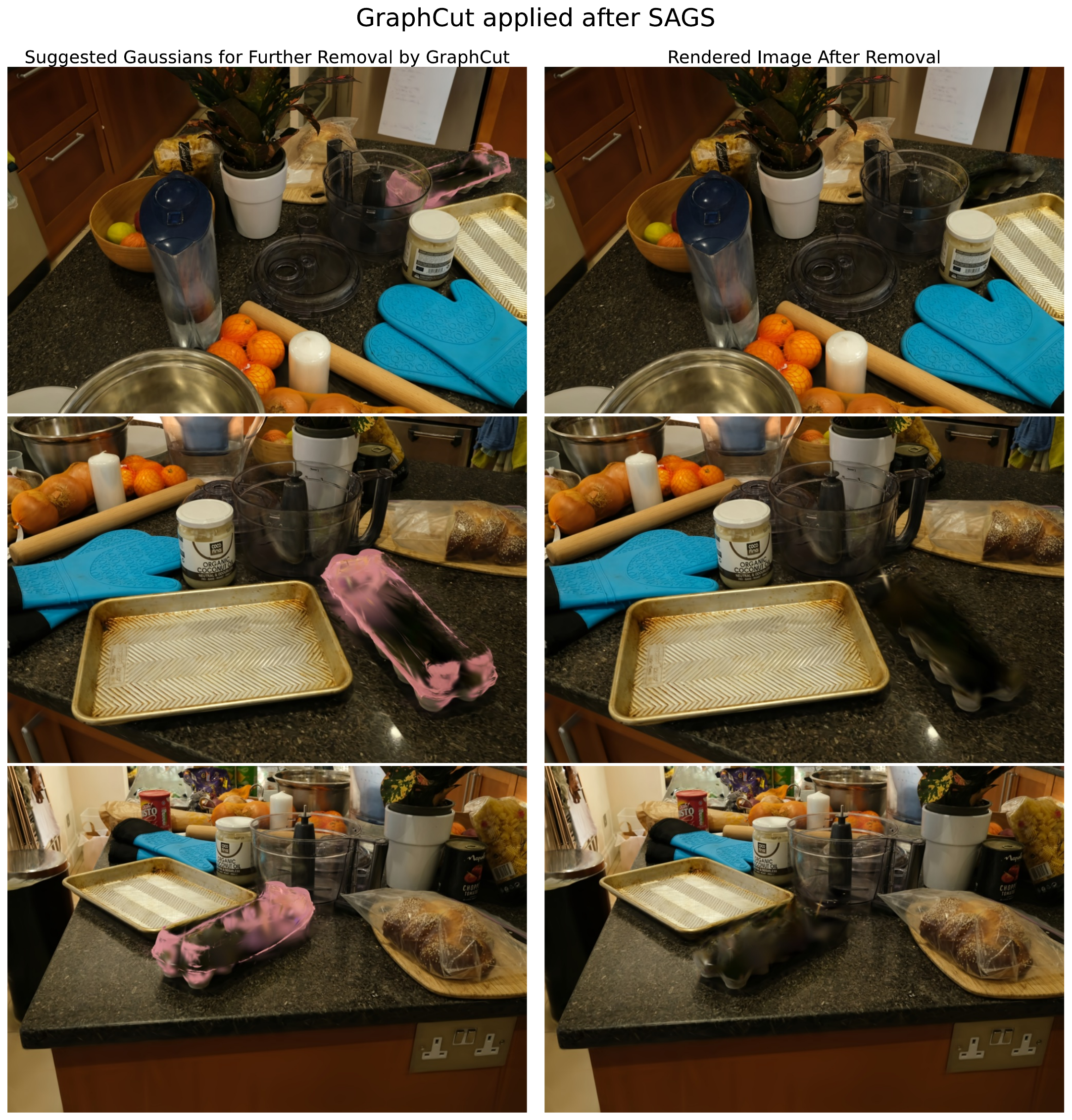}
    \caption{
    \textbf{Graph-based refinement of the removal on SAGS~\cite{hu2024semantic}.}
    Left: the image shows the rendering after removal and before refinement.
    The Gaussians to be removed by the refinement are colored in pink.
    Right: the images are rendered after the refined removal.
    The improved removal of the `Egg-box' in the right image demonstrates the effectiveness of the proposed graph-based refinement in identifying and eliminating residual parts left by the initial method.
    }
    \label{fig:graphcut_sags_counter_eggbox}
\end{figure*}

\begin{figure*}
    \centering
    \includegraphics[width=\linewidth]{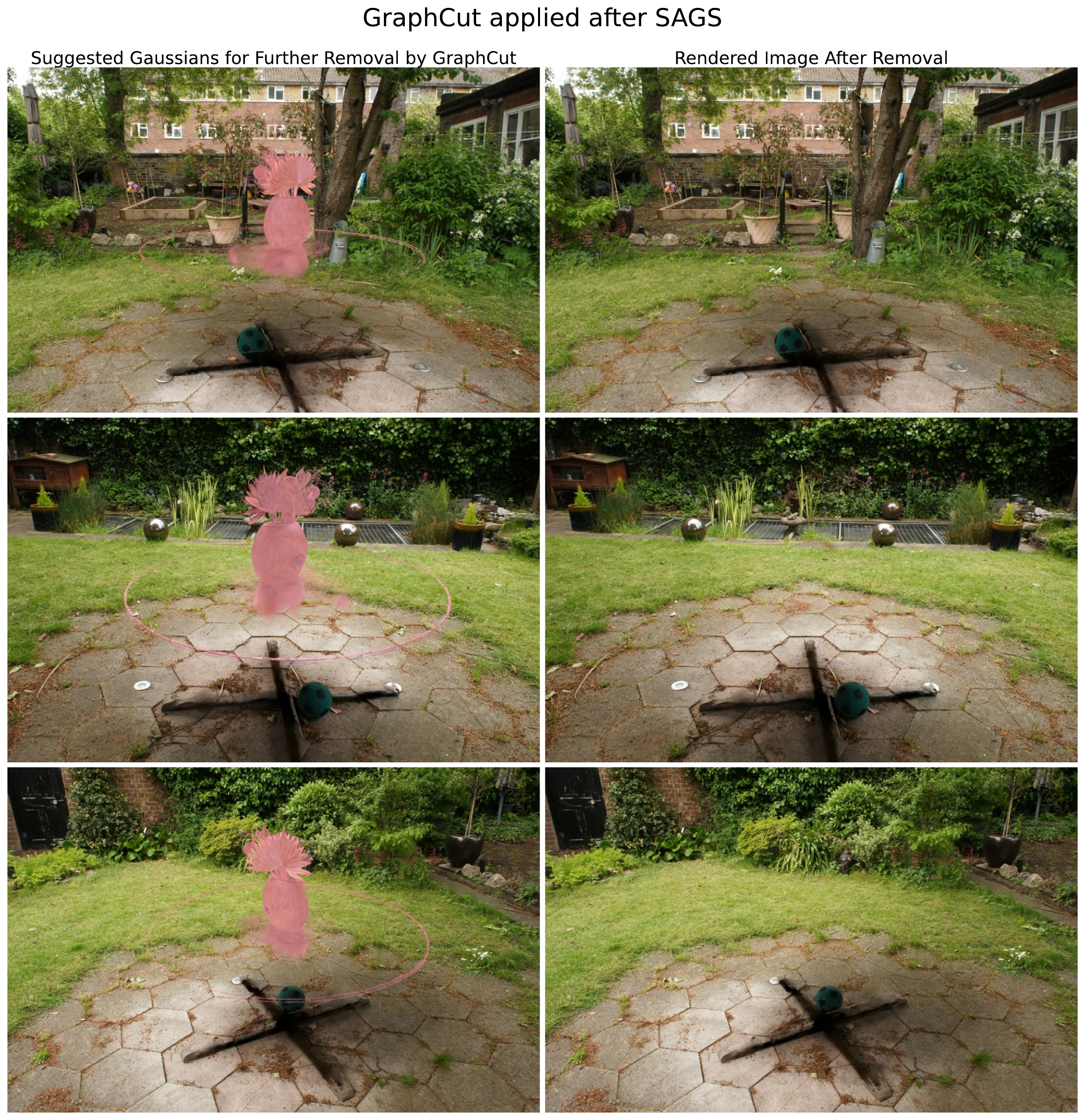}
    \caption{
    \textbf{Graph-based refinement of the removal on SAGS~\cite{hu2024semantic}.}
    Left: the image shows the rendering after removal and before refinement.
    The Gaussians to be removed by the refinement are colored in pink.
    Right: the images are rendered after the refined removal.
    The improved removal of the `Table' in the right image demonstrates the effectiveness of the proposed graph-based refinement in identifying and eliminating residual parts left by the initial method.
    }
    \label{fig:graphcut_sags_garden_table}
\end{figure*}

Failures, marked by "*", occur when the graph can not be initialized, preventing further refinement.
This is the case when the removal method being refined does not output a meaningful set of Gaussians to start from, \eg, FGS~\cite{zhou2024feature} does not remove the `Counter-eggbox' or the `Room-slippers' as indicated by their low IOU$_\text{drop}^{\uparrow}$ at $0.08$ and $0.0$.
The same occurs for SAGS~\cite{hu2024semantic} on the `Counter-plant' and the `Room-coffee table'.
The refinement can also fail outside of this scenario, \eg, FGS~\cite{zhou2024feature} on the `Garden-table'.
Still, the refinement remains relevant as one can decide to include it or not based on whether it improves the metrics.
\cref{fig:graphcut_feature3dgs_room_table}-\ref{fig:graphcut_sags_garden_table} show examples of the Gaussians removed by the refinement.
These results demonstrate that the proposed graph-based refinement can improve the quality of the removal.

\begin{table*}[t]
\centering
  \begin{tabular}{lcc|cc|cc}
    \toprule
      & \multicolumn{2}{c}{IoU$_{\text{drop}}\uparrow$} 
    & \multicolumn{2}{c}{acc$_{\Delta\text{depth}}\uparrow$}  
    & \multicolumn{2}{c}{sim$_{\text{SAM}}\downarrow$}  \\
    \cmidrule(lr){2-7}
    &  FGS & SAGS  &  FGS & SAGS  &  FGS & SAGS \\
        \midrule
    Plant           & 0.75& *& 1.00& *& 0.13& *\\
    Egg Box         & *& 0.60 (+0.04)& *& 0.98 (+0.12)& *& 0.47\\
    Plant-r         & 0.56 (+0.03)& 0.26 (+0.09)& 0.98 (+0.01)& 0.37 (+0.04)& 0.16 (-0.06)& 0.57\\
    Slippers        & *& 0.42 (+0.17)& *& 0.91& *& 0.45\\
    Coffee Table    & 0.57& *& 0.78 (+0.11)& *& 0.12 (-0.14)& *\\
    Truck           & 0.66 (+0.04)& *& 0.96& *& 0.14 (-0.21)& *\\
    Table-g         & *& 0.90 (+0.09)& *& 0.99 (+0.01)& *& 0.04\\
    Vase            & 0.9 (+0.11)& 0.97 (+0.01)& 1.00 (+0.01)& 1.00 (+0.04)& 0.00& 0.11\\
    \bottomrule
  \end{tabular}
  \caption{\textbf{Graph-based Refinement Results.}
  The table reports the performance of the removal after the proposed graph-based refinement on Feature3DGS~\cite{zhou2024feature} and SAGS~\cite{hu2024semantic}.
  Fail cases are reported with $*$ and occur when the graph can not be initialized from the output of the removal methods.
  }
  \label{tab:graphcut_results}
\end{table*}

\subsection{Metrics}

\PAR{Semantic IOU Evaluation.}
\cref{tab:detailed sematic iou} presents a detailed evaluation of IoU values before and after object removal, the IoU difference (IoU$_\text{drop}^\uparrow$), and the percentage reduction from the initial IoU. 
These metrics are categorized by scene, removed object, and removal method. 
Higher IoU$_\text{drop}^\uparrow$ values, ranging from -1 to 1, indicate more effective object removal, as denoted by the $\uparrow$ symbol.

\begin{figure*}
    \centering
    \includegraphics[width=\linewidth]{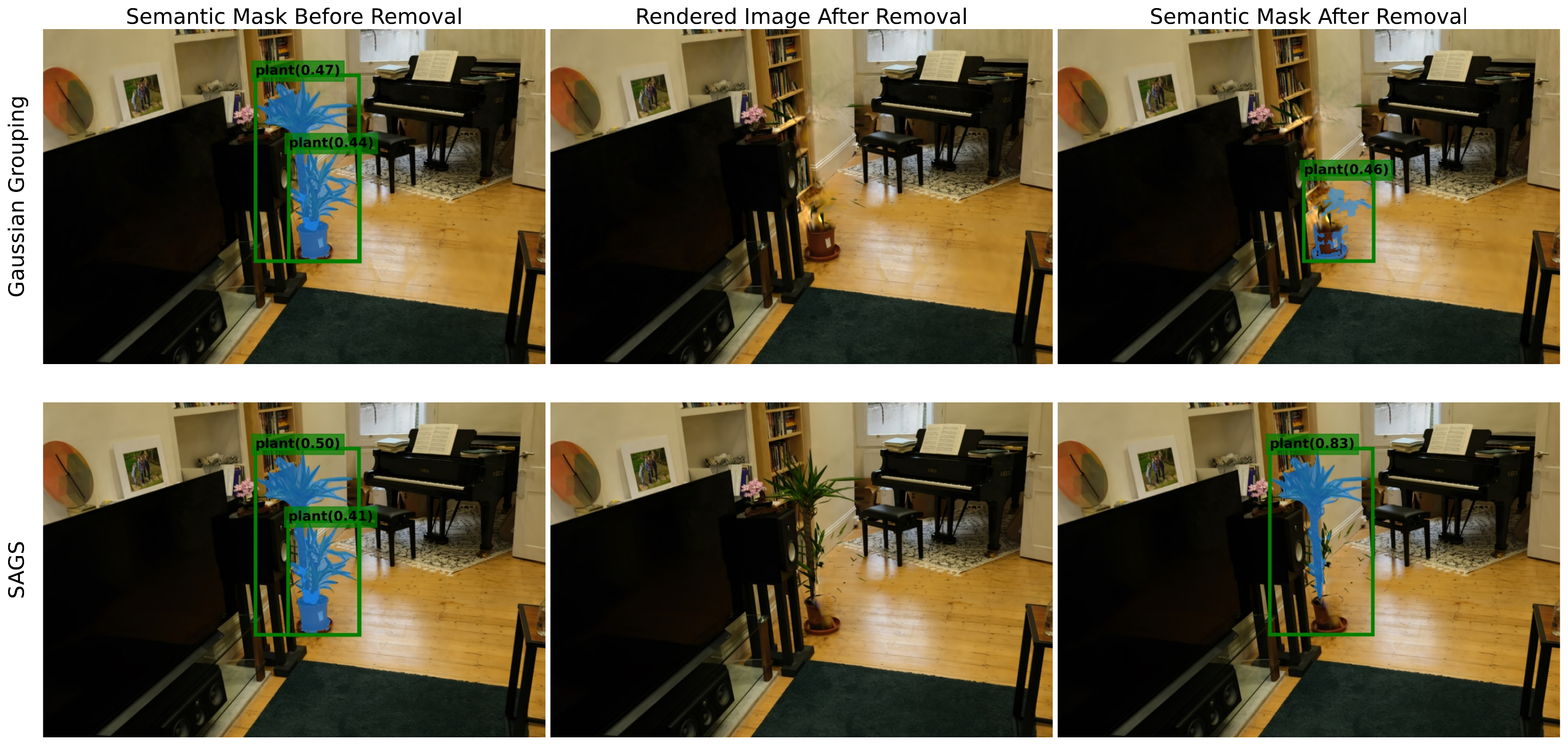}
    \caption{\textbf{GroundedSAM~\cite{kirillov2023segany,liu2023grounding,ren2024grounded} Semantic Segmentation Examples} on renderings from GaussianGrouping~\cite{gaussian_grouping} and SAGS~\cite{hu2024semantic}.
    The left image shows the rendering before removal with its corresponding segmentation. 
    The middle image presents the rendered scene after removal and the right image shows the segmentation after removal.
    Even after removal, GroundedSAM still segments the `Plant', which indicates that the object or parts of the object remain in the scene.
    }
    \label{fig:gsam_room_plant}
\end{figure*}

\begin{figure*}
    \centering
    \includegraphics[width=\linewidth]{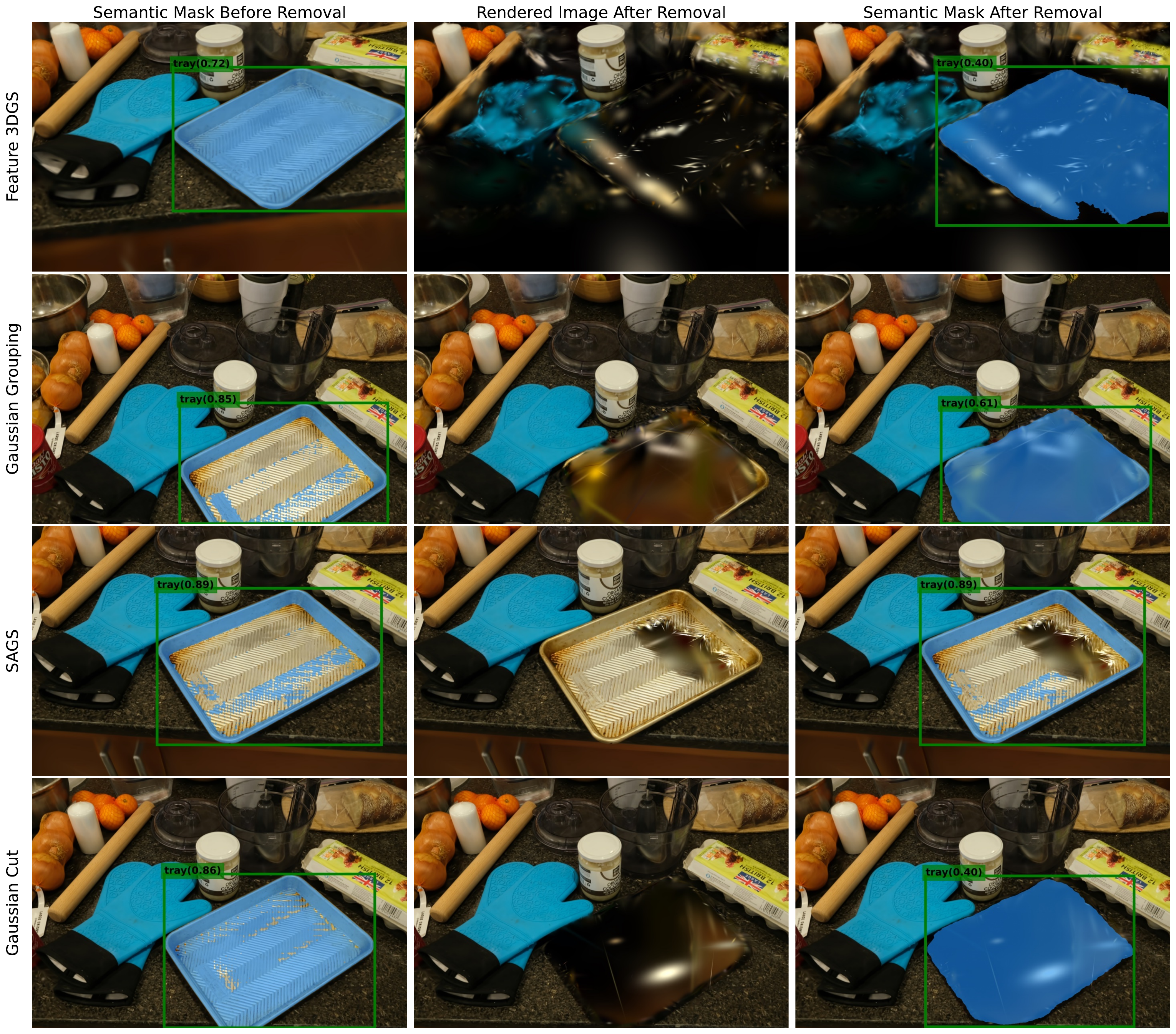}
    \caption{\textbf{GroundedSAM~\cite{kirillov2023segany,liu2023grounding,ren2024grounded} Semantic Segmentation Examples} on renderings from Feature3DGS~\cite{zhou2024feature}, GaussianGrouping~\cite{gaussian_grouping}, SAGS~\cite{hu2024semantic}, and GaussianCut~\cite{jain2024gaussiancut}.
    The left image shows the rendering before removal with its corresponding segmentation. 
    The middle image presents the rendered scene after removal and the right image shows the segmentation after removal. 
    Even after removal, GroundedSAM still segments the `Tray', which indicates either that the object or parts of the object remain in the scene (SAGS) or that there remains invisible information about the object that the segmentation can detect (Feature3DGS, GaussianGrouping, GaussianCut).
    }
    \label{fig:gsam_counter_baking_tray}
\end{figure*}

\cref{tab:sematic_iou} reports the ratio of images where the object remains segmented post removal.
We define that an object is segmented if the semantic segmentation's IoU is higher than a threshold.
We report results on three thresholds: $\{0.5, 0.7, 0.9\}$.
\cref{fig:gsam_room_plant,fig:gsam_counter_baking_tray} provides examples of semantic segmentation before and after removal.

\begin{table*}[t] 
  \centering
  \begin{tabular}{lllccc}
    \toprule
    \textbf{Scene} & \textbf{Object} & \textbf{Method} & mIoU$_\text{pre}$ & mIoU$_\text{post}$ & IoU$_\text{drop}^\uparrow$ \\

    \midrule
    Counter & Baking Tray & Gaussian Grouping   & 0.61 & 0.08& \underline{0.53 / 86.9}\\
            &             & Feature 3DGS        & 0.54 & 0.21& 0.34 / 63.0       \\
            &             & SAGS                & 0.62 & 0.52& 0.10 / 16.1       \\
            &             & Gaussian Cut        & 0.63 & 0.01& \textbf{0.62} / 98.4 \\
    \cmidrule(lr){2-6}& Plant       & Gaussian Grouping   & 0.84& 0.00& \underline{0.84 / 100}\\
            &             & Feature 3DGS        & 0.75& 0.00& 0.75 / 100 \\
            &             & SAGS                & 0.85& 0.82& 0.03 / 3.50\\
            &             & Gaussian Cut        & 0.86& 0.00& \textbf{0.86 / 100}\\
    \cmidrule(lr){2-6}& Blue Gloves & Gaussian Grouping   & 0.75& 0.15& \textbf{0.60 / 80.0}\\
            &             & Feature 3DGS        & 0.67& 0.66& 0.01 / 1.50\\
            &             & SAGS                & 0.74& 0.64& 0.10 / 13.5\\
            &             & Gaussian Cut        & 0.74& 0.15& \textbf{0.60 / 81.1}\\
    \cmidrule(lr){2-6}& Egg Box     & Gaussian Grouping   & 0.63& 0.00& \textbf{0.63 / 100}\\
            &             & Feature 3DGS        & 0.78& 0.7& 0.08 / 10.3\\
            &             & SAGS                & 0.6& 0.04& 0.56 / 93.3\\
            &             & Gaussian Cut        & 0.63& 0.01& \underline{0.62 / 98.4}\\
    \cmidrule(lr){1-6}
    Room& Plant       & Gaussian Grouping   & 0.50& 0.23& 0.26 / 13.0\\
            &             & Feature 3DGS    & 0.53& 0.00& \textbf{0.53 / 100}\\
            &             & SAGS            & 0.52& 0.35& 0.17 / 32.7\\
            &             & Gaussian Cut    & 0.53& 0.00& \textbf{0.53 / 100}\\
    \cmidrule(lr){2-6}& Slippers    & Gaussian Grouping   & 0.96& 0.14& \textbf{0.82 / 85.4}\\
            &             & Feature 3DGS        & 0.96& 0.96& 0.00 / 0.00\\
            &             & SAGS                & 0.96& 0.71& 0.25 / 26.0\\
            &             & Gaussian Cut        & 0.97& 0.48& \underline{0.48 / 49.5}\\
    \cmidrule(lr){2-6}& Coffee Table & Gaussian Grouping & 0.88& 0.02& \textbf{0.86 / 97.7}\\
            &             & Feature 3DGS        & 0.86& 0.29& 0.57 / 66.3\\
            &             & SAGS                & 0.89& 0.89& 0.00 / 0.00\\
            &             & Gaussian Cut         & 0.89& 0.03& \textbf{0.86 / 96.6}\\
    \cmidrule(lr){1-6}
Kitchen& Truck& Gaussian Grouping           & 0.67& 0.06& 0.61 / 91.0\\
            &             & Feature 3DGS    & 0.67& 0.05& 0.62 / 92.5\\
            &             & SAGS            & 0.67& 0.00&  \textbf{0.67 / 100}\\
            &             & Gaussian Cut    & 0.67& 0.01& \underline{0.66 / 98.5}\\
    \cmidrule(lr){1-6}
    Garden& Table& Gaussian Grouping                & 0.89& 0.41& 0.48 / 42.7\\
            &             & Feature 3DGS            & 0.90& 0.24& 0.67 / 74.4\\
            &             & SAGS                    & 0.90& 0.09& \underline{0.81 / 90.0}\\
            &             & Gaussian Cut            & 0.90& 0.04& \textbf{0.86 / 95.6}\\
    \cmidrule(lr){2-6}& Ball& Gaussian Grouping     & 0.16& 0.00& 0.16 / 100\\
            &             & Feature 3DGS            & 0.06& 0.06& 0.00 / 0.00\\
            &             & SAGS                    & 0.41& 0.0& \underline{0.41 / 100}\\
            &             & Gaussian Cut            & 0.42& 0.00& \textbf{0.42 / 100}\\
    \cmidrule(lr){2-6}& Vase& Gaussian Grouping     & 0.85& 0.22& 0.64 / 75.3\\
            &             & Feature 3DGS            & 0.90& 0.11&  0.79 / 87.8 \\
            &             & SAGS                    & 0.97& 0.01& \underline{0.96 / 99.0}\\
            &             & Gaussian Cut            & 0.97& 0.01& \textbf{0.97 / 100}\\
    \bottomrule
  \end{tabular}
  \caption{\textbf{Breakdown of the proposed semantic segmentation IoU$_\text{drop}^\uparrow$ metric.} 
  IoU$_{\text{drop}}$ = IoU$_{\text{post}}$ - IoU$_{\text{pre}}$ and the higher, the better the removal.
  The best-performing method is highlighted in bold and the second-best is underlined.
  We also report the individual segmentation IoUs before and after removal, IoU$_\text{pre}$ and IoU$_\text{post}$ respectively.
  }
  \label{tab:detailed sematic iou}
\end{table*}

\begin{table*}[t] 
  \centering
  \begin{tabular}{lllccc}
  \toprule
    \textbf{Scene} & \textbf{Object} & \textbf{Method} & acc$_{\text{IoU-post} > 0.5}\downarrow$& acc$_{\text{IoU-post} > 0.7}\downarrow$& acc$_{\text{IoU-post} > 0.9}\downarrow$\\
    \midrule
    Counter & Baking Tray & Gaussian Grouping & 8.50& 8.50&      5.70\\
            &             & Feature 3DGS     & 21.7& 19.8&      17.9\\
            &             & SAGS          & 51.9& 44.3&      30.2\\
            &             & Gaussian Cut           & 0.09& 0.09&      0.09\\
    \cmidrule(lr){2-6}
            & Plant      & Gaussian Grouping & 0.00& 0.00&      0.00\\
            &             & Feature 3DGS     & 0.00& 0.00&      0.00\\
            &             & SAGS          & 83.3& 83.3&      83.3\\
            &             & Gaussian Cut           & 0.00& 0.00&      0.00\\
    \cmidrule(lr){2-6}
            & Blue Gloves & Gaussian Grouping & 16.3& 16.3&      9.60\\
            &             & Feature 3DGS     & 72.1& 66.3&      52.9\\
            &             & SAGS          & 66.3& 50.0&      44.2\\
            &             & Gaussian Cut           & 17.3& 17.3&      3.8\\
    \cmidrule(lr){2-6}
            & Egg Box     & Gaussian Grouping  & 0.00& 0.00&      0.00\\
            &             & Feature 3DGS     & 80.4& 79.4&      71.0\\
            &             & SAGS          & 4.10& 4.10&      3.10\\
            &             & Gaussian Cut           & 1.00& 0.00&      0.00\\
    \cmidrule(lr){1-6}
    Room& Plant       & Gaussian Grouping & 20.0& 12.0&      8.00\\
            &             & Feature 3DGS      & 0.00& 0.00&      0.00\\
            &             & SAGS          & 28.0& 20.0&      8.0\%\\
            &             & Gaussian Cut           & 0.00& 0.00&      0.00\\
    \cmidrule(lr){2-6}
            & Slippers    & Gaussian Grouping & 14.7& 14.7&      13.2\\
            &             & Feature 3DGS     & 98.5& 98.6&      98.5\\
            &             & SAGS          & 72.1& 67.6&      22.1\\
            &             & Gaussian Cut           & 55.9& 41.2&      16.2\\
    \cmidrule(lr){2-6}
            & Coffee Table & Gaussian Grouping & 1.00& 1.00&      1.00\\
            &             & Feature 3DGS     & 38.4& 21.2&      5.10\\
            &             & SAGS          & 90.9& 89.9&      89.9\\
            &             & Gaussian Cut           & 1.00& 1.00&      1.00\\
    \cmidrule(lr){1-6}
Kitchen& Truck& Gaussian Grouping & 7.80& 1.50&      0.00\\
            &             & Feature 3DGS     & 4.90& 4.90&      4.40\\
            &             & SAGS           & 0.00& 0.00&      0.00\\
            &             & Gaussian Cut           & 0.05& 0.00&      0.00\\
    \cmidrule(lr){1-6}
    Garden& Table& Gaussian Grouping & 45.9& 29.0&      10.0\\
            &             & Feature 3DGS     & 29.7& 18.2&      0.00\\
            &             & SAGS          & 12.2& 7.40&      0.00\\
            &             & Gaussian Cut           & 4.70& 3.30&      0.00\\
    \cmidrule(lr){2-6}& Ball& Gaussian Grouping  & 0.00& 0.00&      0.00\\
            &             & Feature 3DGS     & 6.10& 6.10&      6.10\\
            &             & SAGS           & 0.00& 0.00&      0.00\\
            &             & Gaussian Cut            & 0.00& 0.00&      0.00\\
    \cmidrule(lr){2-6}& Vase& Gaussian Grouping & 21.6& 20.2&      19.6\\
            &             & Feature 3DGS     & 10.8& 9.50&      7.40\\
            &             & SAGS           & 0.00& 0.00&      0.00\\
            &             & Gaussian Cut            & 0.00& 0.00&      0.00\\
    \bottomrule
  \end{tabular}
  \caption{\textbf{Additional evaluation based on how well the object is segmentated after removal.}
  This table presents the ratio of images where the object is still segmented even after removal.
  We define that the object is segmented if the semantic segmentation IoU is higher than a threshold.
  We report the results for the thresholds \{0.5, 0.7, and 0.9\}.
  }
  \label{tab:sematic_iou}
\end{table*}

\begin{figure*}
    \centering
    \includegraphics[width=\linewidth]{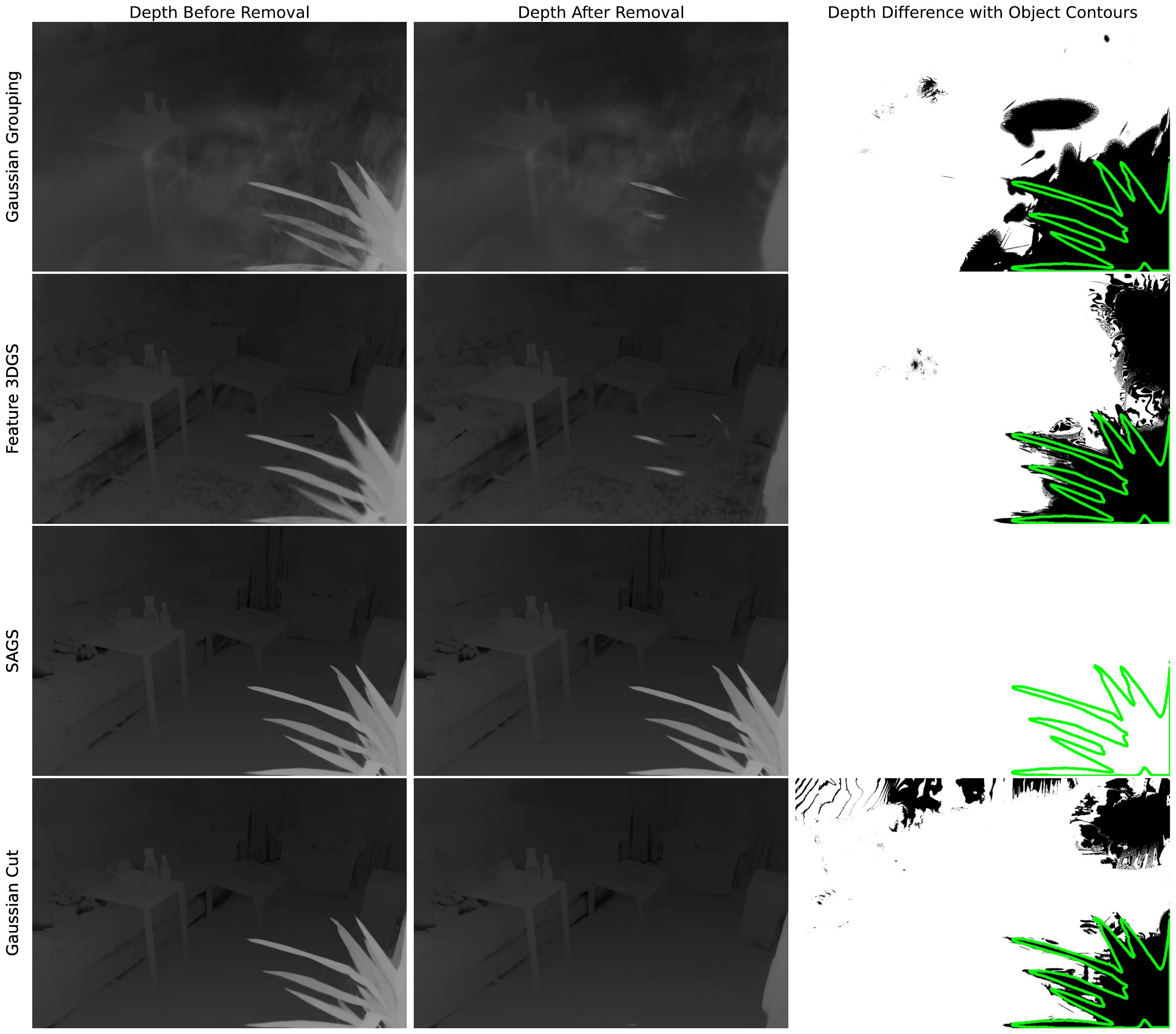}
    \caption{
    \textbf{Depth rendering before and after object removal} on Feature3DGS~\cite{zhou2024feature}, GaussianGrouping~\cite{gaussian_grouping}, SAGS~\cite{hu2024semantic}, and GaussianCut~\cite{jain2024gaussiancut}.
    Left-Right: the original depth map, depth after removal, and the difference map highlighting the removed object’s contours. 
    Notably, SAGS~\cite{hu2024semantic} did not capture any depth difference. 
    While other methods exhibit depth changes (black region in the right image), the changes do not overlay with the pseudo-ground-truth object mask (green outline), which suggests that the removal is subpar.
    }
    \label{fig:depth_diff_room_plant}
\end{figure*}

\begin{figure*}
    \centering
    \includegraphics[width=\linewidth]{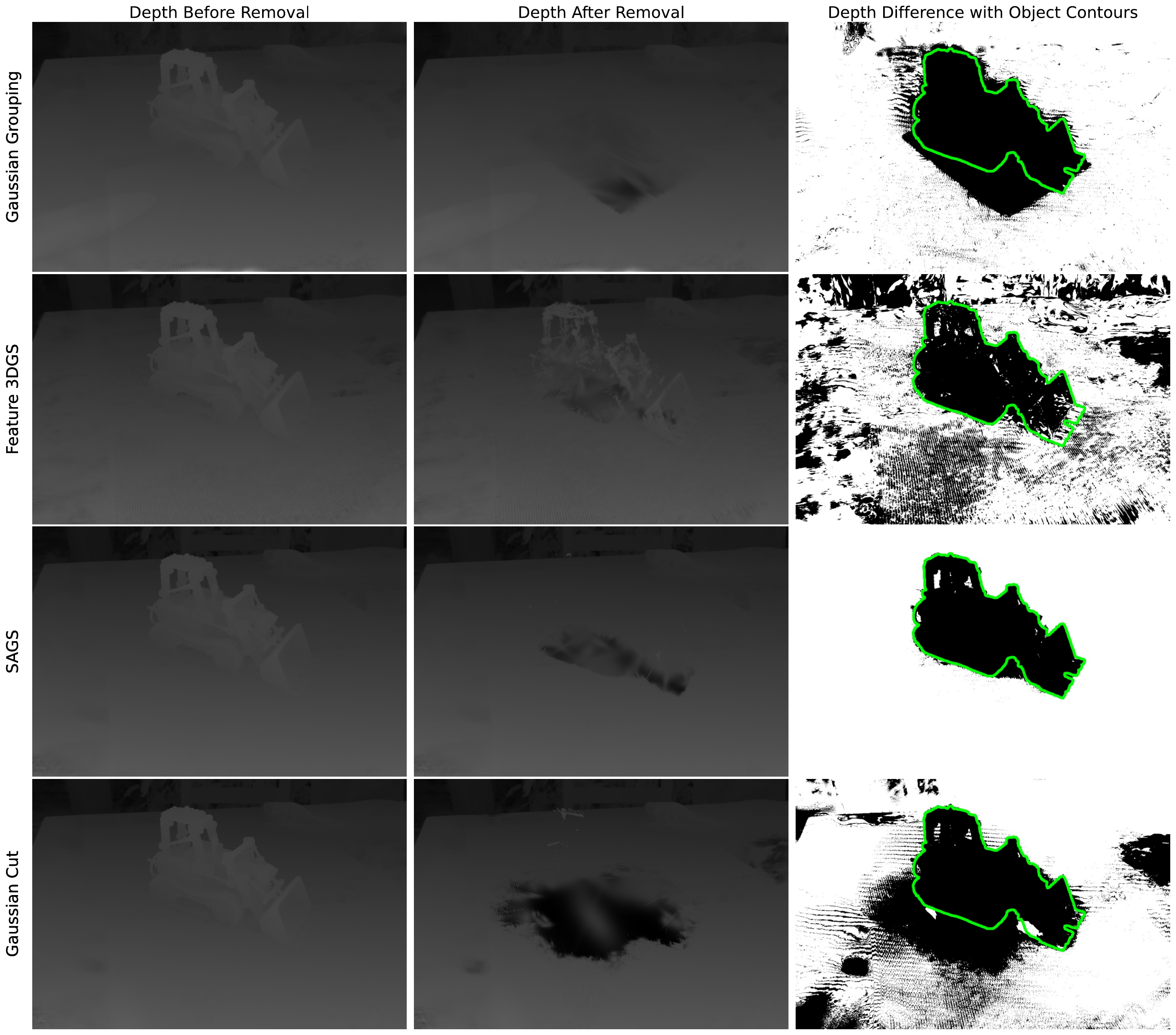}
    \caption{
    \textbf{Depth rendering before and after object removal} on Feature3DGS~\cite{zhou2024feature}, GaussianGrouping~\cite{gaussian_grouping}, SAGS~\cite{hu2024semantic}, and GaussianCut~\cite{jain2024gaussiancut}.
    Left-Right: the original depth map, depth after removal, and the difference map highlighting the removed object's contours.
    SAGS achieves the best removal as measured by the depth differences that overlay well with the pseudo-ground-truth object mask (green outline).
    In contrast, other methods induce depth changes beyond the object’s boundaries, which indicates unintended modifications of the scene's representation.
    }
    \label{fig:depth_diff_kitchen_truck}
\end{figure*}

\PAR{Depth difference after removal.}
~\cref{fig:depth_diff_room_plant} and ~\cref{fig:depth_diff_kitchen_truck} illustrate the effect of object removal on the depth map.

\clearpage
\clearpage

{
    \small
    \bibliographystyle{ieeenat_fullname}
    \bibliography{main}
}

\end{document}

%% file: preamble.tex
\newcommand{\B}[1]{{{\textbf{#1}}}}
\newcommand{\U}[1]{{{\underline{#1}}}}

\usepackage{comment}
\usepackage[acronym]{glossaries}
\usepackage{multirow}

\newcommand{\PAR}[1]{\noindent{\bf #1~}}

\makeglossaries{}
\newacronym{sota}{SotA}{State-of-the-Art}
\newacronym{nn}{NN}{Nearest-Neighbor}
\newacronym{sfm}{SfM}{Structure-from-Motion}
\newacronym{slam}{SLAM}{Simultaneous Localization And Mapping}
\newacronym{tab}{Tab.}{Table}
\newacronym{sdf}{SDF}{Signed-Distance-Function}
\newacronym{udf}{UDF}{Unsigned-Distance-Function}
\newacronym{gs}{GS}{Gaussian Splats}
\newacronym{iou}{IoU}{Intersection over Union}
\newacronym{sam}{SAM}{SegmentAnything}
\newacronym{nerf}{NeRF}{Neural Radiance Field}

\newcommand{\sam}{\gls{sam}~}
